\documentclass{article}
\pdfoutput=1
\usepackage{microtype}
\usepackage{graphicx}
\usepackage{subfigure}
\usepackage{xspace}
\usepackage{amsmath}
\usepackage{enumitem}
\usepackage{booktabs} 
\usepackage{caption}
\usepackage{xcolor}
\usepackage{multirow}

\usepackage{iclr2021_conference,times}
\iclrfinalcopy

\usepackage[utf8]{inputenc} 
\usepackage[T1]{fontenc}    
\usepackage{hyperref}       
\usepackage{url}            
\usepackage{booktabs}       
\usepackage{amsfonts}       
\usepackage{nicefrac}       
\usepackage{microtype}      

\def\method{Quant-Noise\@\xspace}

\title{Training with Quantization Noise for \\ Extreme Model Compression}

\author{Angela Fan \thanks{Equal contribution. Corresponding authors: \texttt{angelafan@fb.com}, \texttt{pstock@fb.com}} \\
  Facebook AI Research, LORIA\\
    \And
  Pierre Stock \footnotemark[1]~~\thanks{Univ Lyon, Inria, CNRS, ENS de Lyon, UCB Lyon 1, LIP UMR 5668, F-69342, Lyon, France} \\
  Facebook AI Research, Inria\\
     \And
  Benjamin Graham \\
  Facebook AI Research\\
     \And
  Edouard Grave \\
  Facebook AI Research\\
     \And
  Rémi Gribonval \footnotemark[2] \\
  Inria\\
     \And
  Hervé Jégou \\
  Facebook AI Research\\
     \And
  Armand Joulin \\
  Facebook AI Research\\
}

\begin{document}

\maketitle

\begin{abstract}
We tackle the problem of producing compact models, maximizing their accuracy for a given model size. 
A standard solution is to train networks with Quantization Aware Training~\citep{jacob2018quantization}, where the weights are quantized during training and the gradients approximated with the Straight-Through Estimator~\citep{bengio2013estimating}. 
In this paper, we extend this approach to work beyond \texttt{int8} fixed-point quantization with extreme compression methods where the approximations introduced by STE are severe, such as Product Quantization. 
Our proposal is to only quantize a different random subset of weights during each forward, allowing for unbiased gradients to flow through the other weights. 
Controlling the amount of noise and its form  allows for extreme compression rates while maintaining the performance of the original model. 
As a result we establish new state-of-the-art compromises between accuracy and model size both in natural language processing and image classification. 
For example, applying our method to state-of-the-art Transformer and ConvNet architectures, we can achieve 82.5\% accuracy on MNLI by compressing RoBERTa to 14\,MB and 80.0\% top-1 accuracy on ImageNet by compressing an EfficientNet-B3 to 3.3\,MB.\footnote{Code available at \url{https://github.com/pytorch/fairseq/tree/master/examples/quant_noise}}
\end{abstract}



\section{Introduction}

Many of the best performing neural network architectures in real-world applications have a large number of parameters.
For example, the current standard machine translation architecture, Transformer~\citep{vaswani2017attention}, has layers that contain millions of parameters. 
Even models that are designed to jointly optimize the performance and the parameter efficiency, such as EfficientNets~\citep{tan2019efficientnet}, 
still require dozens to hundreds of megabytes, which limits their applications to domains like robotics or virtual assistants.

Model compression schemes reduce the memory footprint of overparametrized models. Pruning ~\citep{lecun1990optimal} and distillation~\citep{hinton2015distilling} remove parameters by reducing the number of network weights. 
In contrast, quantization focuses on reducing the bits per weight. This makes quantization particularly interesting when compressing models that have already been carefully optimized in terms of network architecture. Whereas deleting weights or whole hidden units will inevitably lead to a drop in performance, we demonstrate that quantizing the weights can be performed with little to no loss in accuracy.

Popular postprocessing quantization methods, like scalar quantization, replace the floating-point weights of a trained network by a lower-precision representation, like fixed-width integers~\citep{vanhoucke2011improving}. 
These approaches achieve a good compression rate with the additional benefit of accelerating inference on supporting hardware.  
However, the errors made by these approximations accumulate in the computations operated during the forward pass, inducing a significant drop in performance~\citep{DBLP:journals/corr/abs-1907-05686}.

A solution to address this drifting effect 
is to directly quantize the network during training.
This raises two challenges. 
First, the discretization operators have a null gradient --- the derivative with respect to the input is zero almost everywhere.
This requires special workarounds to train a network with these operators. 
The second challenge that often comes with these workarounds is the discrepancy that appears between the train and test functions implemented by the network. 
Quantization Aware Training (QAT)~\citep{jacob2018quantization} resolves these issues by quantizing all the weights during the forward and using a straight through estimator (STE)~\citep{bengio2013estimating} to compute the gradient.
This works when the error introduced by STE is small, like with \texttt{int8} quantization, but does not suffice in compression regimes where the approximation made by the compression is more severe.

\begin{figure}[t]
\begin{center} 
\centerline{\includegraphics[width=\textwidth]{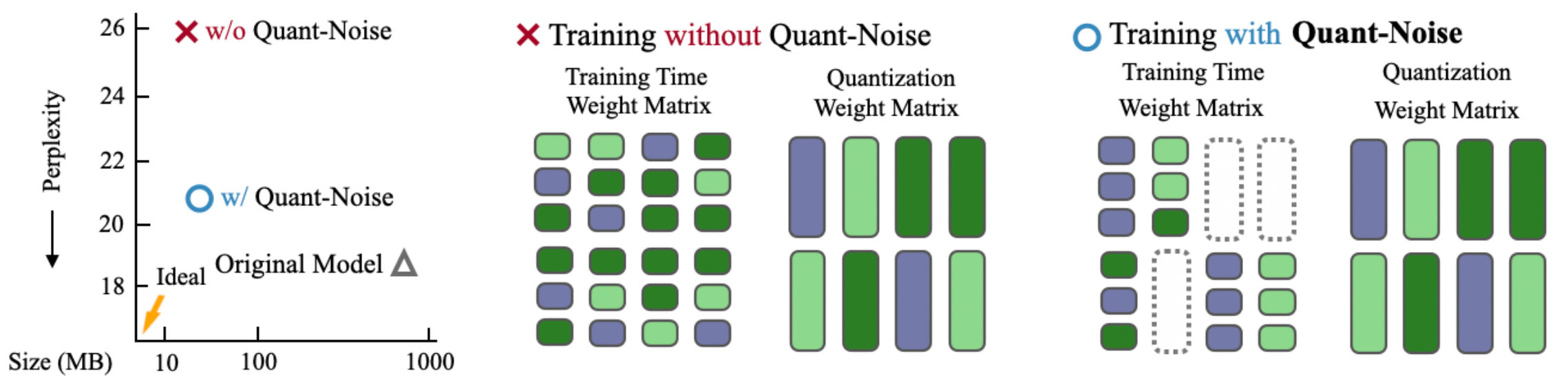}}
\caption{\textbf{\method} trains models to be resilient to inference-time quantization by mimicking the effect of the quantization method during training time. This allows for extreme compression rates without much loss in accuracy on a variety of tasks and benchmarks.
}
\label{fig:pullfigure}
\end{center}
\vskip -0.38in
\end{figure}

In this work, we show that quantizing only a subset of weights instead of the entire network during training is more stable for high compression schemes.
Indeed, by quantizing only a random fraction of the network at each forward, most the weights are updated with unbiased gradients.
Interestingly, we show that our method can employ a simpler quantization scheme during the training.
This is particularly useful for quantizers with trainable parameters, such as Product Quantizer (PQ), for which our quantization proxy is not parametrized.
Our approach simply applies a quantization noise, called \method, to a random subset of the weights, see Figure~\ref{fig:pullfigure}. 
We observe that this makes a network resilient to various types of discretization methods: it significantly improves the accuracy associated with (a) low precision representation of weights like \texttt{int8}; and (b) state-of-the-art PQ.  Further, we demonstrate that \method can be applied to existing trained networks as a post-processing step, to improve the performance network after quantization. 

In summary, this paper makes the following contributions: 
\begin{itemize}
    \item  We introduce the \method technique to learn networks that are more resilient to a variety of quantization methods such as \texttt{int4}, \texttt{int8}, and PQ;
    \item Adding Quant-Noise to PQ leads to new state-of-the-art trade-offs between accuracy and model size. For instance, for natural language processing (NLP), we reach 82.5\% accuracy on MNLI by compressing RoBERTa to 14\,MB. Similarly for computer vision, we report 80.0\% top-1 accuracy on ImageNet by compressing an EfficientNet-B3 to 3.3\,MB;
    \item By combining PQ and \texttt{int8} to quantize weights and activations for networks trained with \method, we obtain extreme compression with fixed-precision computation and achieve 79.8\% top-1 accuracy on ImageNet and $21.1$ perplexity on WikiText-103.
\end{itemize}

\section{Related Work}
\label{sec:related}

\paragraph{Model compression.}
Many compression methods focus on efficient parameterization, via weight pruning \citep{lecun1990optimal,li2016pruning,huang2018condensenet,mittal2018recovering}, weight sharing \citep{dehghani2018universal, turc2019well, lan2019albert} or with dedicated architectures~\citep{tan2019efficientnet,DBLP:journals/corr/ZhangZLS17,howard2019searching}.
Weight pruning is implemented during training~\citep{louizos2017learning} or as a fine-tuning post-processing step~\citep{han2015learning,han2015deep}. 
Many pruning methods are unstructured, i.e., remove individual weights~\citep{lecun1990optimal,molchanov2017variational}.
On the other hand, structured pruning methods follow the structure of the weights 
to reduce both the memory footprint and the inference time of a model~\citep{li2016pruning,luo2017thinet, fan2019reducing}.
We refer the reader to~\cite{liu2018rethinking} for a  review of different pruning strategies.

Other authors have worked on lightweight architectures, by modifying existing models~\citep{zhang2018accelerating,wu2018pay,sukhbaatar2019adaptive} or developing new networks, such as MobileNet \citep{howard2019searching}, ShuffleNet \citep{DBLP:journals/corr/ZhangZLS17}, and EfficientNet \citep{tan2019efficientnet} in vision. 

Finally, knowledge distillation~\citep{hinton2015distilling} has been applied to sentence representation~\citep{turc2019well,distilbert,Sun_2019,zhao2019extreme, jiao2019tinybert}, to reduce the size of a BERT model~\citep{devlin2018bert}.

\paragraph{Quantization.}
There are extensive studies of scalar quantization to train networks with low-precision weights and activations~\citep{DBLP:journals/corr/CourbariauxBD15, DBLP:journals/corr/CourbariauxB16, rastegari2016xnor, mcdonnell2018training}. 
These methods benefit from specialized hardware to also improve the runtime during inference~\citep{vanhoucke2011improving}.
Other quantization methods such as Vector Quantization (VQ) and PQ~\citep{Jegou:2011:PQN:1916487.1916695} quantize blocks of weights simultaneously to achieve higher compression rate~\citep{DBLP:journals/corr/abs-1907-05686,gong2014compressing,joulin2016fasttext,carreiraperpin2017model}. 
Closer to our work, several works have focused at simultaneously training and quantizing a network~\citep{jacob2018quantization,krishnamoorthi2018quantizing,gupta2015deep,dong2019stochastic}.
\cite{gupta2015deep} assigns weights to a quantized bin stochastically which is specific to scalar quantization, but allows training with fixed point arithmetic.
Finally, our method can be interpreted as a form of Bayesian compression~\citep{louizos2017learning}, using the Bayesian interpretation of Dropout~\citep{gal2016dropout}.
As opposed to their work, we select our noise to match the weight transformation of a target quantization methods without restricting it to a scale mixture prior.

\section{Quantizing Neural Networks}
\label{sec:preliminaries}

In this section, we present the principles of quantization, several standard quantization methods, and describe how to combine scalar and product quantization.
For clarity, we focus on the case of a fixed real matrix~$\mathbf W \in \mathbf R^{n \times p}$. 
We suppose that this matrix is split into~$m\times q$ blocks~$\mathbf{b}_{kl}$:
\begin{equation}
\mathbf W = 
\begin{pmatrix}
    \mathbf b_{11}& \cdots & \mathbf b_{1q}\\
    \vdots      & \ddots & \vdots \\
    \mathbf b_{m1} & \cdots & \mathbf b_{mq}
\end{pmatrix},
\label{eq:matrix}
\end{equation}
where the nature of these blocks is determined by the quantization method.
A codebook is a set of~$K$ vectors, i.e.,~$\mathcal{C}= \{\mathbf c[1],\dots,\mathbf c[K]\}$. 
Quantization methods compress the matrix~$\mathbf W$ by 
assigning to each block~$\mathbf b_{kl}$ an index that points to a codeword~$\mathbf c$ in a codebook~$\mathcal{C}$, and storing the codebook~$\mathcal{C}$ and the resulting indices (as the entries~$\mathbf I_{kl}$ of an index matrix~$\mathbf I$) instead of the real weights.
During the inference, they reconstruct an approximation $\mathbf{ \widehat{W}}$ of the original matrix $\mathbf W$ such that $\mathbf{ \widehat{\mathbf b}}_{kl} = \mathbf c[\mathbf I_{kl}]$.

We distinguish scalar quantization, such as \texttt{int8}, where each block $\mathbf{b}_{kl}$ consists of a single weight, from vector quantization, where several weights are quantized jointly.

\subsection{Fixed-point Scalar Quantization}

Fixed-point scalar quantization methods replace floating-point number representations by low-precision fixed-point representations.
They simultaneously reduce a model's memory footprint and accelerate inference by using fixed-point arithmetic on supporting hardware.

Fixed-point scalar quantization operates on blocks that represent a single weight, i.e., $\mathbf b_{kl} = \mathbf W_{kl}$. 
Floating-point weights are replaced by $N$ bit fixed-point numbers~\citep{gupta2015deep}, with the extreme case of binarization where $N=1$~\citep{DBLP:journals/corr/CourbariauxBD15}.
More precisely, the weights are rounded to one of~$2^N$ possible codewords. 
These codewords correspond to bins evenly spaced by a scale factor~$s$ and shifted by a bias~$z$.
Each weight~$\mathbf W_{kl}$ is mapped to its nearest codeword~$c$ by successively quantizing with $z\mapsto \text{round}(\mathbf W_{kl} / s + z)$ and dequantizing with the inverse operation:
\begin{equation}
\label{eq:int8}
\mathbf c=\left(\text{round}(\mathbf W_{kl} / s + z)  - z\right)\times s,
\end{equation}
where we compute the scale and bias as:
\begin{equation*}
 s=\frac{\max\mathbf W-\min\mathbf W}{2^N - 1}  \text{~~~and~~~}  z=\text{round}(\min\mathbf W / s). 
\end{equation*}
We focus on this uniform rounding scheme instead of other non-uniform schemes~\citep{choi2018pact,li2019additive}, because it allows for fixed-point arithmetic with  implementations in PyTorch and Tensorflow (see Appendix).
The compression rate is~$\times 32/N$.
The activations are also rounded to $N$-bit fixed-point numbers. With \texttt{int8} for instance, this leads to $\times 2$ to $\times 4$ faster inference on dedicated hardware.
In this work, we consider both \texttt{int4} and \texttt{int8} quantization.

\subsection{Product Quantization}
Several quantization methods work on groups of weights, such as vectors, to benefit from the correlation induced by the structure of the network. 
In this work, we focus on Product Quantization for its good performance at extreme compression ratio~\citep{DBLP:journals/corr/abs-1907-05686}.

\paragraph{Traditional PQ.} In vector quantization methods, the blocks are predefined groups of weights instead of single weights. 
The codewords are groups of values, and the index matrix $\mathbf{I}$ maps groups of weights from the matrix $\mathbf W$ to these codewords.
In this section, we present the Product Quantization framework as it generalizes both scalar and vector quantization. We consider the case where we apply PQ to the \emph{columns} of $\mathbf W$ and thus assume that $q = p$.
 
Traditional vector quantization techniques split the matrix $\mathbf W$ into its $p$ columns and learn a codebook on the resulting $p$ vectors. 
Instead, Product Quantization splits each column into $m$ subvectors and learns the same codebook for each of the resulting $m \times p$ subvectors. 
Each quantized vector is subsequently obtained by assigning its subvectors to the nearest codeword in the codebook.
Learning the codebook is traditionally done using $k$-means with a fixed number $K$ of centroids, typically $K=256$ to store the index matrix $\mathbf I$ using \texttt{int8}. 
Thus, the objective function is written as:
\begin{equation}
    \| \mathbf W - \mathbf{\widehat W}\|_2^2 = \sum_{k,l} \|\mathbf b_{kl} - \mathbf c[\mathbf I_{kl}]\|_2^2.
\end{equation}
PQ shares representations between subvectors, which allows for higher compression rates than $\texttt{intN}$. 

\paragraph{Iterative PQ.}
When quantizing a full network rather than a single matrix, extreme compression with PQ induces a quantization drift as reconstruction error accumulates ~\citep{DBLP:journals/corr/abs-1907-05686}. Indeed, subsequent layers take as input the output of preceding layers, which are modified by the quantization of the preceding layers.
This creates a drift in the network activations, resulting in large losses of performance.
A solution proposed by~\cite{DBLP:journals/corr/abs-1907-05686}, which we call \textbf{iterative PQ} (iPQ), is to quantize layers sequentially from the lowest to the highest, and finetune the upper layers as the lower layers are quantized, under the supervision of the uncompressed (teacher) model. 
Codewords of each layer are finetuned by averaging the gradients of their assigned elements with gradient steps:
 \begin{equation}
     \mathbf c \leftarrow \mathbf c - \eta \frac{1}{| J_{\mathbf c}|}\sum_{(k, l) \in  J_{\mathbf c}} \frac{\partial \mathcal L}{\partial \mathbf b_{kl}},
 \end{equation}
where $ J_\mathbf c = \{(k, l) \mid \mathbf c[\mathbf I_{kl}] = \mathbf c\}$, $\mathcal L$ is the loss function and $\eta>0$ is a learning rate.
This adapts the upper layers to the drift appearing in their inputs, reducing the impact of the quantization approximation on the overall performance.

\subsection{Combining Fixed-Point with Product Quantization}\label{subsec:mix}

Fixed-point quantization and Product Quantization are often regarded as competing choices, but can be advantageously combined. 
Indeed, PQ/iPQ compresses the network by replacing vectors of weights by their assigned centroids, but these centroids are in floating-point precision.
Fixed-point quantization compresses both activations and weights to fixed-point representations.
Combining both approaches means that the vectors of weights are mapped to centroids that are compressed to fixed-point representations, along with the activations.
This benefits from the extreme compression ratio of iPQ and the finite-precision arithmetics of \texttt{intN} quantization.

More precisely, for a given matrix, we store the \texttt{int8} representation of the $K$ centroids of dimension $d$ along with the $\log_2 K$ representations of the centroid assignments of the $m\times p$ subvectors. 
The \texttt{int8} representation of the centroids is obtained with Eq.~(\ref{eq:int8}).
The overall storage of the matrix and activations during a forward pass with batch size $1$ (recalling that the input dimension is n) writes
\begin{equation}
    M = 8 \times K d +  \log_2K \times m p + 8 \times n \text{~bits.}
\end{equation}
In particular, when $K=256$, the centroid assignments are also stored in \texttt{int8}, which means that every value required for a forward pass is stored in an \texttt{int8} format.
We divide by $4$ the \texttt{float32} overhead of storing the centroids, although the storage requirement associated with the centroids is small compared to the cost of indexing the subvectors for standard networks. 
In contrast to iPQ alone where we only quantize the weights, we also quantize the activations using \texttt{int8}.
We evaluate this approach on both natural language processing and computer vision tasks in Section~\ref{sec:results}.

\section{Method}
\label{sec:method}

Deep networks are not exposed to the noise caused by the quantization drift during training, leading to suboptimal performance.
A solution to make the network robust to quantization is to introduce it during training. 
Quantization Aware Training~(QAT)~\citep{jacob2018quantization} exposes the network during training by quantizing weights during the forward pass.
This transformation is not differentiable and gradients are approximated with a straight through estimator (STE)~\citep{bengio2013estimating,DBLP:journals/corr/CourbariauxB16}.
STE introduces a bias in the gradients that depends on level of quantization of the weights, and thus, the compression ratio.
In this section, we propose a simple modification to control this induced bias with a stochastic amelioration of QAT, called \method.
The idea is to quantize a randomly selected fraction of the weights instead of the full network as in QAT, leaving some unbiased gradients flow through unquantized weights.
Our general formulation can simulate the effect of both quantization and of pruning during training. 
 
\subsection{Training Networks with Quantization Noise}

We consider the case of a real matrix $\mathbf{W}$ as in Section~\ref{sec:preliminaries}. 
During the training of a network, our proposed \method method works as follows:
first, we compute blocks $\mathbf b_{kl}$ related to a target quantization method.
Then, during each forward pass, we randomly select a subset of these blocks and apply some distortion to them.

More formally, given a set of tuples of indices $J \subset \{(k, l)\}$ for $1 \leq k \leq m$, $1\leq l \leq q$ and a \emph{distortion} or \emph{noise} function $\varphi$ acting on a block, we define an operator $\psi(\cdot \mid J)$ such that, for each block $\mathbf b_{kl}$, we apply the following transformation:
\begin{equation}
  \psi(\mathbf b_{kl}\mid J) = 
  \begin{cases} 
      \varphi(\mathbf b_{kl}) & \text{if } (k,l) \in J, \\
      \mathbf b_{kl} & \text{otherwise.}
   \end{cases}
\end{equation}
The noise function $\varphi$ simulates the change in the weights produced by the target quantization method (see Section~\ref{subsec:varphi} for details).
We replace the matrix $\mathbf W$ by the resulting noisy matrix $\mathbf W_{\text{noise}}$ during the forward pass to compute a noisy output $\mathbf{y}_{\text{noise}}$, i.e.,
\begin{equation}
    \mathbf W_{\text{noise}} = \left(\psi(\mathbf b_{kl}\mid J)\right)_{kl} ~~~~\text{~and~}~~~~ \mathbf{y}_{\text{noise}} =   \mathbf{x}\mathbf W_{\text{noise}}
\end{equation}
where $\mathbf{x}$ is an input vector.
During the backward pass, we apply STE, which amounts to replacing the distorted weights $\mathbf W_{\text{noise}}$ by their non-distorted counterparts.
Note that our approach is equivalent to QAT when $J$ containts all the tuples of indices.
However, an advantage of \method over QAT is that unbiased gradients continue to flow via blocks unaffected by the noise.
As these blocks are randomly selected for each forward, we guarantee that each weight regularly sees gradients that are not affected by the nature of the function $\varphi$.
As a side effect, our quantization noise regularizes the network in a similar way as DropConnect~\citep{wan2013regularization} or LayerDrop~\citep{fan2019reducing}.

\paragraph{Composing quantization noises.}
As noise operators are compositionally commutative, we can make a network robust to a combination of quantization methods by composing their noise operators:
\begin{equation}
\label{eq:comp}
\psi(\mathbf b_{kl}~|~J) = \psi_{1} \circ \psi_{2} (\mathbf b_{kl}~|~J).
\end{equation}
This property is particularly useful to combine quantization with pruning operators during training, as well as combining scalar quantization with product quantization.

\subsection{Adding Noise to Specific Quantization Methods}\label{subsec:varphi}
In this section, we propose several implementations of the noise function $\varphi$ for the quantization methods described in Section~\ref{sec:preliminaries}.
We also show how to handle pruning with it.

\paragraph{Fixed-point scalar quantization.} 
In \texttt{intN} quantization, the blocks are atomic and weights are rounded to their nearest neighbor in the codebook.
The function $\varphi$ replaces weight $\mathbf W_{kl}$ with the output of the rounding function defined in Eq.~(\ref{eq:int8}), i.e.,
\begin{equation}
\varphi_{\texttt{intN}}(w) = \left(\text{round}(w / s + z)  - z\right)\times s, 
\end{equation}
where $s$ and $z$ are updated during training.
In particular, the application of \method to \texttt{int8} scalar quantization is a stochastic amelioration of QAT.

\begin{table*}[t]
    \small 
    \centering
    \begin{tabular}{l@{\hspace{20pt}} c ccc cl@{\hspace{20pt}} ccc}
    \toprule
 \bf Quantization Scheme &~~& \multicolumn{3}{c}{\bf Language Modeling} &~~& \multicolumn{3}{c}{\bf Image Classification}\\
 && \multicolumn{3}{c}{\small 16-layer Transformer} &~~& \multicolumn{3}{c}{\small EfficientNet-B3}\\
 && \multicolumn{3}{c}{\small Wikitext-103} &~~& \multicolumn{3}{c}{\small ImageNet-1k}\\
\cmidrule{3-5}
\cmidrule{7-9}
&& Size & Compression & PPL && Size & Compression & Top-1 \\
    \midrule
Uncompressed model && $942$ & $\times\ \ 1$ & $18.3$ && $46.7$ & $\times\ \ 1$ & $81.5$ \\
    \midrule
    \texttt{int4} quantization & & $118$ & $\times\ \ 8$ & $39.4$ && $5.8$ & $\times\ \ 8$ & $45.3$\\ 
    - trained with QAT & & $118$ & $\times\ \ 8$ & 34.1 && $5.8$ & $\times\ \ 8$  &  $59.4$\\ 
    - trained with \method & & $118$ & $\times\ \ 8$ & $\mathbf{21.8}$ && $5.8$ & $\times\ \ 8$  & $\mathbf{67.8}$ \\ 
    \midrule
    \texttt{int8} quantization & & $236$ & $\times\ \ 4$ & $19.6$ && $11.7$ & $\times\ \ 4$ & $80.7$\\ 
- trained with QAT & & $236$ & $\times\ \ 4$ & $21.0$ && $11.7$ & $\times\ \ 4$ & $80.8$\\ 
- trained with \method & &  $236$ & $\times\ \ 4$ & $\mathbf{18.7}$ && $11.7$ & $\times\ \ 4$ & $\mathbf{80.9}$\\ 
    \midrule
    iPQ &  & $38$ & $\times\ \ 25$ & $25.2$ & &  $3.3$ & $\times\ \ 14$ & $79.0$   \\ 
- trained with QAT &&  $38$ & $\times\ \ 25$ & $41.2$ & & $3.3$ &  $\times\ \ 14$ &  $55.7$  \\ 
- trained with \method && $38$ & $\times\ \ 25$ & $\mathbf{20.7}$ & & $3.3$ &  $\times\ \ 14$ & $\mathbf{80.0}$ \\ 
    \midrule 
    \midrule 
    iPQ \& \texttt{int8} + \method & &  $38$ & $\times\ \ 25$ & $21.1$ && $3.1$ & $\times \ \ 15$ & $79.8$ \\
    \bottomrule
    \end{tabular}
 \caption{\small \textbf{Comparison of different quantization schemes with and without \method} on language modeling and image classification. For language modeling, we train a Transformer on the Wikitext-103 benchmark and report perplexity (PPL) on test. For image classification, we train a EfficientNet-B3 on the ImageNet-1k benchmark and report top-1 accuracy on validation and use our re-implementation of EfficientNet-B3. The original implementation of~\cite{tan2019efficientnet} achieves an uncompressed Top-1 accuracy of $81.9\%$. For both settings, we report model size in megabyte (MB) and the compression ratio compared to the original model.}
    \label{tab:quantization_comparison}
\end{table*}

\paragraph{Product quantization.}
\label{sect:ipq_application}
As opposed to \texttt{intN}, codebooks in PQ require a clustering step based on weight values.
During training, we learn codewords online and use the resulting centroids to implement the quantization noise.
More precisely, the noise function $\varphi_{\text{PQ}}$ assigns a selected block $\mathbf b$ to its nearest codeword in the associated codebook $\mathcal{C}$:
\begin{equation}
 \varphi_{\text{PQ}} (\mathbf v) = \text{argmin}_{\mathbf c\in \mathcal{C}} \|\mathbf b - \mathbf c\|_2^2.
\end{equation}
Updating the codebooks online works well. 
However, empirically, running $k$-means once per epoch is faster and does not noticeably modify the resulting accuracy.

Note that computing the exact noise function for PQ is computationally demanding.
We propose a simpler and faster alternative approximation $\varphi_{\text{proxy}}$ to the operational transformation of PQ and iPQ.
The noise function simply zeroes out the subvectors of the selected blocks, i.e., 
$
\varphi_{\text{proxy}}(\mathbf v) = 0.
$
As a sidenote, we considered other alternatives, for instance one where the subvectors are mapped to the mean subvector. 
In practice, we found that these approximations lead to similar performance, see Section~\ref{sec:ablations}.
This proxy noise function is a form of Structured Dropout and encourages correlations between the subvectors.
This correlation is beneficial to the subsequent clustering involved in PQ/iPQ. 

\paragraph{Adding pruning to the quantization noise.}
The specific form of quantization noise can be adjusted to incorporate additional noise specific to pruning.
We simply combine the noise operators of quantization and pruning by composing them following Eq.~(\ref{eq:comp}).
We consider the pruning noise function of~\cite{fan2019reducing} where they randomly drop predefined structures during training.
In particular, we focus on \emph{LayerDrop}, where the structures are the residual blocks of highway-like layers~\citep{srivastava2015highway}, as most modern architectures, such as ResNet or Transformer, are composed of this structure.
More precisely, the corresponding noise operator over residual blocks $\mathbf{v}$ is 
$
\varphi_{\text{LayerDrop}}(\mathbf v) = 0.
$
For pruning, we do not use STE to backpropagate the gradient of pruned weights, as dropping them entirely during training has the  benefit of speeding convergence~\citep{huang2016deep}. 
Once a model is trained with LayerDrop, the number of layers kept at inference can be adapted to match computation budget or time constraint.


\section{Results}
\label{sec:results}

We demonstrate the impact of \method on the performance of several quantization schemes in a variety of settings (see Appendix - Sec.~\ref{sec:experiments}).

\begin{figure}[t]
    \includegraphics[width=.32\textwidth]{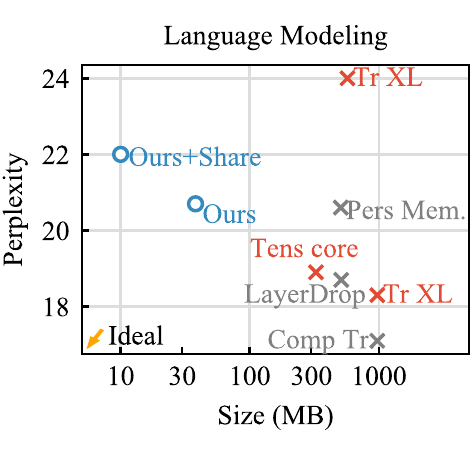} \hfill
    \includegraphics[width=.32\textwidth]{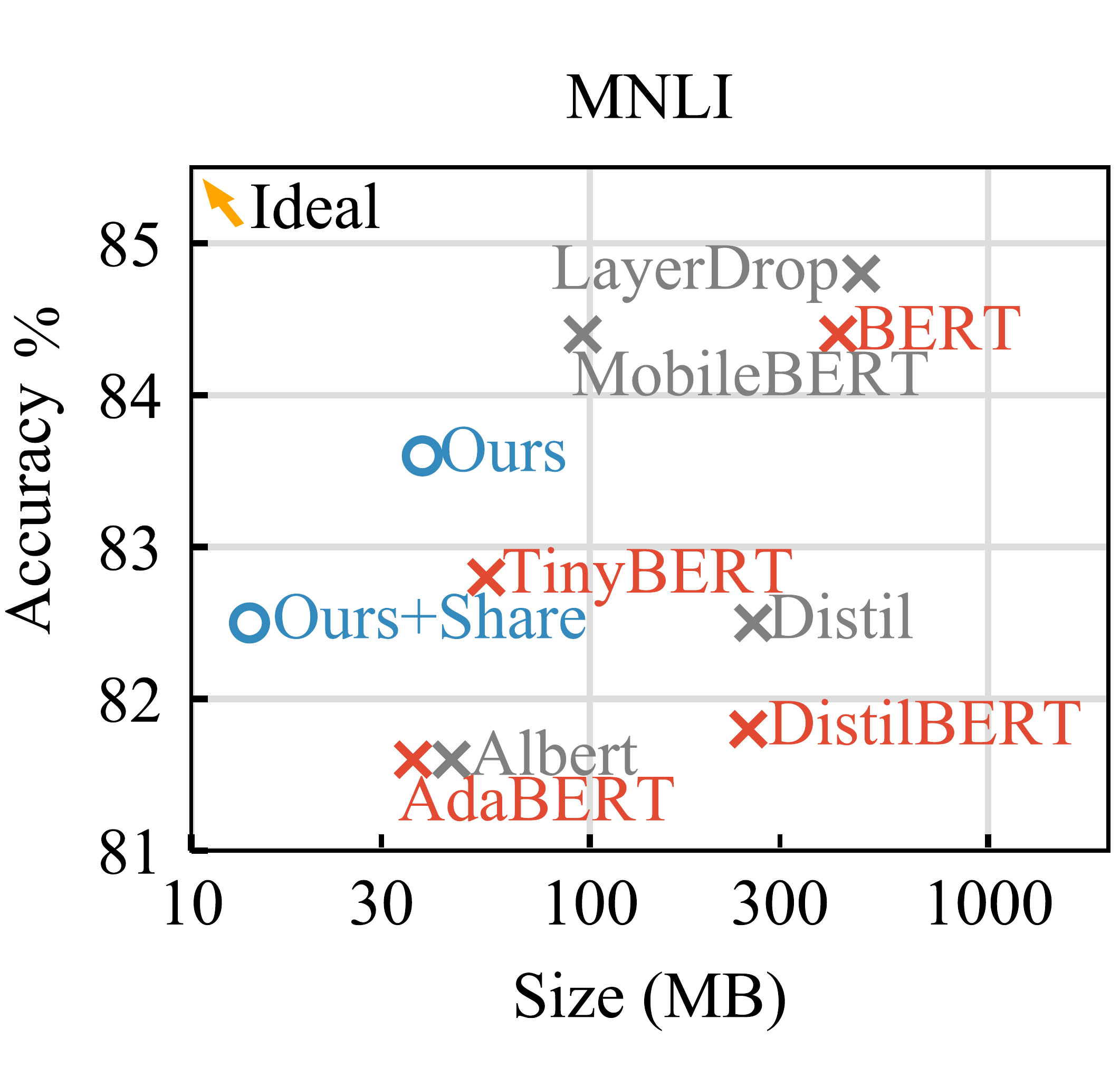} \hfill 
    \includegraphics[width=.32\textwidth]{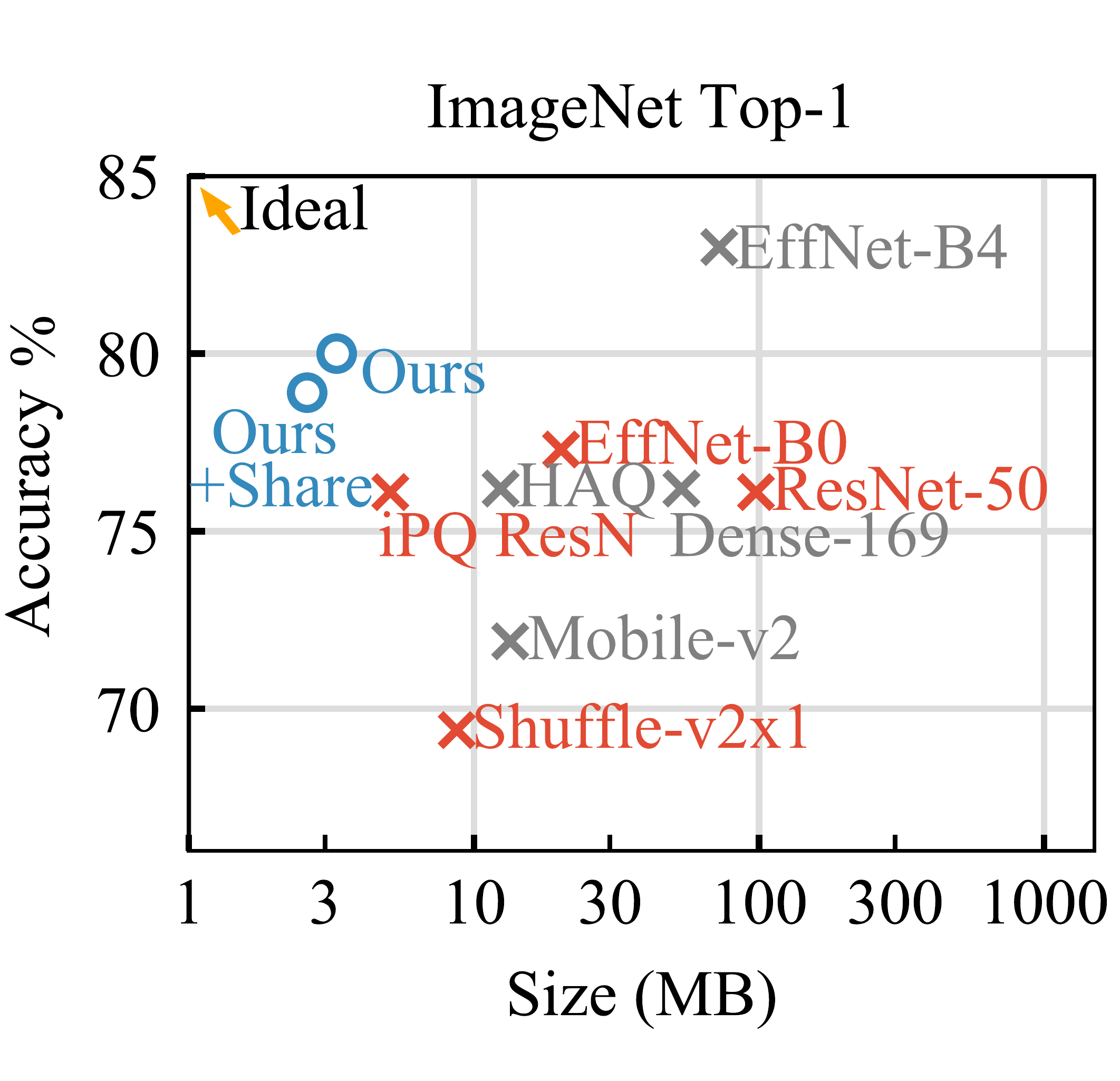}
    \vspace{-13pt}
    \caption{\small \textbf{Performance as a function of model size.}
We compare models quantized with PQ and trained with the related \method to the state of the art.
\textbf{(a)} Test perplexity on Wikitext-103 \textbf{(b)} Dev Accuracy on MNLI \textbf{(c)} ImageNet Top-1 accuracy. Model size is shown in megabytes on a log scale. Red and gray coloring indicates existing work, with different colors for visual distinction.
    \label{fig:main_result}}
        \vspace{-5pt}
\end{figure}

\begin{table}[t]
\setlength{\tabcolsep}{5.5pt}
\centering
\small 
    \begin{tabular}{l c ccc c ccc c ccc}
    \toprule
    &~~& \multicolumn{3}{c}{\bf Language modeling}  	&~~& \multicolumn{3}{c}{\bf Sentence Representation} &~~& \multicolumn{3}{c}{\bf Image Classification}\\
\cmidrule{3-5}\cmidrule{7-9}\cmidrule{11-13}
    	      &&  Comp. &  Size &  PPL		&&  Comp. &  Size &  Acc.		&&  Comp. &  Size &  Acc.	\\
    \midrule
    \multicolumn{4}{l}{\small\emph{Unquantized models}}	\\
    Original model && $\times\ \ 1$ & $942$ & $18.3$ 	&& $\times\ \ 1$ & $480$ &   $84.8$ 	&& $\times\  1$ & $46.7$ & $81.5$ 	\\ 
    + Sharing  	   && $\times\ \ 1.8$ & $510$ & $18.7$ 	&& $\times\ \ 1.9$ & $250$ & $84.0$	&& $\times\ \ 1.4 $ & $34.2$ & $80.1$ 	\\ 
    + Pruning 	   && $\times\ \ 3.7$ & $255$ & $22.5$ 	&& $\times\ \ 3.8$ & $125$ & $81.3$	&& $\times\ \ 1.6$ & $29.5$ & $78.5$ 	\\ 
    \midrule                                                                                     
    \multicolumn{4}{l}{\small\emph{Quantized models}}\\                                          
    iPQ 	&& $\times\ \ 24.8$   & $38$ & $25.2$ && $\times\ \ 12.6$ & $38$ & $82.5$	&& $\times\ \ 14.1$ & $3.3$ & $79.0$ \\
    + \method 		&& $\times\ \ 24.8$   & $38$ & $20.7$ && $\times\ \ 12.6$ & $38$ & $83.6$	&& $\times\ \ 14.1$ & $3.3$ & $80.0$  \\ 
    + Sharing 		&& $\times\ \ 49.5$ & $19$ & $22.0$   && $\times\ \ 34.3 $& $14$ & $82.5$ && $\times\ \ 18$  &  $2.6$ & $78.9$  \\ 
    + Pruning 		&& $\times\ \ 94.2$   & $10$ & $24.7$ && $\times\ \ 58.5 $& $8$ & $78.8$ 	&& $\times\ \ 20$  &  $2.3$ & $77.8$   \\ 
    \bottomrule
    \end{tabular}
    \vspace{-7pt}
    \caption{\small \textbf{Decomposing the impact of the different compression schemes.} 
\textbf{(a)} we  train Transformers with Adaptive Input and LayerDrop on Wikitext-103 \textbf{(b)} we pre-train RoBERTA base models with LayerDrop and then finetune on MNLI 
\textbf{(c)} we train an  EfficientNet-B3 on ImageNet. We report the compression ratio w.r.t. to the original model (``comp.'') and the resulting size in MB.
    \label{tab:lm_quant}}
    \vspace{-10pt}
\end{table}

\subsection{Improving Compression with \method}

\method is a regularization method that makes networks more robust to the target quantization scheme or combination of quantization schemes during training.
We show the impact of \method in Table 1 for a variety of quantization methods: \texttt{int8/int4} and iPQ.

We experiment in $2$ different settings: a Transformer network trained for language modeling on WikiText-103 and a EfficientNet-B3 convolutional network trained for image classification on ImageNet-1k.
Our quantization noise framework is general and flexible --- \method improves the performance of quantized models for every quantization scheme in both experimental settings. 
Importantly, \method only changes model training by adding a regularization noise similar to dropout, with no impact on convergence and very limited impact on training speed (< 5\% slower). 

This comparison of different quantization schemes shows that \method works particularly well with high performance quantization methods, like iPQ, where QAT tends to degrade the performances, even compared to quantizing as a post-processing step.
In subsequent experiments in this section, we focus on applications with iPQ because it offers the best trade-off between model performance and compression, and has little negative impact on FLOPS. 

\paragraph{Fixed-Point Product Quantization.} Combining iPQ and \texttt{int8} as described in Section~\ref{subsec:mix} allows us to take advantage of the high compression rate of iPQ with a fixed-point representation of both centroids and activations. As shown in Table~\ref{tab:quantization_comparison}, this combination incurs little loss in accuracy with respect to iPQ + \method. Most of the memory footprint of iPQ comes from indexing and not storing centroids, so the compression ratios are comparable.

\paragraph{Complementarity with Weight Pruning and Sharing.}

We analyze how \method is compatible and complementary with pruning (``+Prune'') and weight sharing (``+Share''), see Appendix for details on weight sharing.
We report results for Language modeling on WikiText-103, pre-trained sentence representations on MNLI and object classification on ImageNet-1k in Table~\ref{tab:lm_quant}. 
The conclusions are remarkably consistent across tasks and benchmarks: \method gives a large improvement over strong iPQ baselines. Combining it with sharing and pruning offers additional interesting operating points of performance vs size. 

\subsection{Comparison with the state of the art} 

We now compare our approach on the same tasks against the state of the art. 
We compare iPQ + \method with 6 methods of network compression for Language modeling, 8 state-of-the-art methods for Text classification, and 8 recent methods evaluate image classification on Imagenet with compressed models. These comparisons demonstrate that \method leads to extreme compression rates at a reasonable cost in accuracy. 
We apply our best quantization setup on competitive models and reduce their memory footprint by  $\times 20-94$ when combining with weight sharing and pruning, offering extreme compression for good performance.

\paragraph{Natural Language Processing.}  
In Figure~\ref{fig:main_result}, we examine the trade-off between performance and model size.
Our quantized RoBERTa offers a competitive trade-off between size and performance with memory reduction methods dedicated to BERT, like TinyBERT, MobileBERT, or AdaBERT.

\paragraph{Image Classification. }
We compress EfficientNet-B3 from $46.7$Mb to $3.3$Mb ($\times 14$ compression) while maintaining high top-1 accuracy ($78.5\%$ versus $80\%$ for the original model). 
As shown in Figure~\ref{fig:main_result}, our quantized EfficientNet-B3 is smaller and more accurate than architectures dedicated to optimize on-device performance with limited size like MobileNet or ShuffleNet. We further evaluate the beneficial effect of Quant-Noise on ResNet-50 to compare directly with \cite{DBLP:journals/corr/abs-1907-05686}. Results shown in Table~\ref{tab:resnet} indicate improvement with Quant-Noise compared to previous work. 

Incorporating pruning noise into quantization is also beneficial. For example, with pruning iPQ+\method reduces size by $\times 25$ with only a drop of $2.4$ PPL in language modeling.
Further, pruning reduces FLOPS by the same ratio as its compression factor, in our case, $\times 2$.
By adding sharing with pruning, in language modeling, we achieve an extreme compression ratio of $\times 94$ with a drop of $6.4$ PPL with FLOPS reduction from pruning entire shared chunks of layers.
For comparison, our $10$ MB model has the same performance as the $570$ MB Transformer-XL base.


\begin{table}[t]
    \centering
    \small 
    \begin{tabular}{lcc c lcc }
    \toprule
    \bf Language Modeling && \bf PPL &~~& \bf RoBERTa && \bf Acc. 	\\ 
    \midrule 
    Train without \method  && 25.2 && Train without \method && 82.5 \\ 
    + Finetune with \method  && 20.9 && + Finetune with \method && 83.4	\\ 
    \midrule 
    Train with \method && 20.7 && Train  with \method && 83.6\\ 
    \bottomrule
    \end{tabular}
    \vspace{-7pt}
    \caption{\small \textbf{\method: Finetuning vs training.}
We report performance after iPQ quantization.
We train with the $\phi_{\text{proxy}}$ noise and finetune with \method, and use it during the transfer to MNLI for each RoBERTa model.
}
    \label{tab:finetune}
\end{table}

\subsection{Finetuning with \method for Post-Processing Quantization}

We explore taking existing models and post-processing with \method instead of training from scratch. 
For language modeling, we train for 10 additional epochs. 
For RoBERTa, we train for 25k additional updates.
Finetuning with \method incorporates the benefits and almost matches training from scratch (Table~\ref{tab:finetune}). In language modeling, there is only a $0.2$ PPL difference. 
We further examine how to incorporate \method more flexibly into pretraining RoBERTa. 
We take an already trained RoBERTa model and incorporate \method during sentence classification finetuning. 
This is effective at compressing while retaining accuracy after quantization.

\section{Conclusion}
\label{sec:conclusion} 
We show that quantizing a random subset of weights during training maintains performance in the high quantization regime.
We validate that \method works with a variety of different quantization schemes on several applications in text and vision. 
Our method can be applied to a combination of iPQ and \texttt{int8} to benefit from extreme compression ratio and fixed-point arithmetic. Finally, we show that \method can be used as a post-processing step to prepare already trained networks for subsequent quantization, to improve the performance of the compressed model.




\bibliography{egbib}

\begin{thebibliography}{77}
\providecommand{\natexlab}[1]{#1}
\providecommand{\url}[1]{\texttt{#1}}
\expandafter\ifx\csname urlstyle\endcsname\relax
  \providecommand{\doi}[1]{doi: #1}\else
  \providecommand{\doi}{doi: \begingroup \urlstyle{rm}\Url}\fi

\bibitem[Adcock et~al.(2019)Adcock, Reis, Singh, Yan, van~der Maaten, Zhang,
  Motwani, Guerin, Goyal, Misra, Gustafson, Changhan, and
  Goyal]{adcock2019classy}
A.~Adcock, V.~Reis, M.~Singh, Z.~Yan, L.~van~der Maaten, K.~Zhang, S.~Motwani,
  J.~Guerin, N.~Goyal, I.~Misra, L.~Gustafson, C.~Changhan, and P.~Goyal.
\newblock Classy vision.
\newblock 2019.

\bibitem[Baevski \& Auli(2018)Baevski and Auli]{baevski2018adaptive}
Alexei Baevski and Michael Auli.
\newblock Adaptive input representations for neural language modeling.
\newblock \emph{arXiv preprint arXiv:1809.10853}, 2018.

\bibitem[Bengio et~al.(2013)Bengio, L{\'e}onard, and
  Courville]{bengio2013estimating}
Yoshua Bengio, Nicholas L{\'e}onard, and Aaron Courville.
\newblock Estimating or propagating gradients through stochastic neurons for
  conditional computation.
\newblock \emph{arXiv preprint arXiv:1308.3432}, 2013.

\bibitem[Bradbury et~al.(2016)Bradbury, Merity, Xiong, and
  Socher]{bradbury2016quasi}
James Bradbury, Stephen Merity, Caiming Xiong, and Richard Socher.
\newblock Quasi-recurrent neural networks.
\newblock \emph{arXiv preprint arXiv:1611.01576}, 2016.

\bibitem[Cao et~al.()Cao, Trivedi, Balasubramanian, and
  Balasubramanian]{decompbert}
Qingqing Cao, Harsh Trivedi, Aruna Balasubramanian, and Niranjan
  Balasubramanian.
\newblock Faster and just as accurate: A simple decomposition for transformer
  models.

\bibitem[Carreira-Perpiñán \& Idelbayev(2017)Carreira-Perpiñán and
  Idelbayev]{carreiraperpin2017model}
Miguel~A. Carreira-Perpiñán and Yerlan Idelbayev.
\newblock Model compression as constrained optimization, with application to
  neural nets. part ii: quantization, 2017.

\bibitem[Chen et~al.(2020)Chen, Li, Qiu, Wang, Li, Ding, Deng, Huang, Lin, and
  Zhou]{chen2020adabert}
Daoyuan Chen, Yaliang Li, Minghui Qiu, Zhen Wang, Bofang Li, Bolin Ding, Hongbo
  Deng, Jun Huang, Wei Lin, and Jingren Zhou.
\newblock Adabert: Task-adaptive bert compression with differentiable neural
  architecture search.
\newblock \emph{arXiv preprint arXiv:2001.04246}, 2020.

\bibitem[Choi et~al.(2018)Choi, Wang, Venkataramani, Chuang, Srinivasan, and
  Gopalakrishnan]{choi2018pact}
Jungwook Choi, Zhuo Wang, Swagath Venkataramani, Pierce I-Jen Chuang,
  Vijayalakshmi Srinivasan, and Kailash Gopalakrishnan.
\newblock Pact: Parameterized clipping activation for quantized neural
  networks.
\newblock \emph{arXiv preprint arXiv:1805.06085}, 2018.

\bibitem[Courbariaux \& Bengio(2016)Courbariaux and
  Bengio]{DBLP:journals/corr/CourbariauxB16}
Matthieu Courbariaux and Yoshua Bengio.
\newblock Binarynet: Training deep neural networks with weights and activations
  constrained to +1 or -1.
\newblock \emph{CoRR}, 2016.

\bibitem[Courbariaux et~al.(2015)Courbariaux, Bengio, and
  David]{DBLP:journals/corr/CourbariauxBD15}
Matthieu Courbariaux, Yoshua Bengio, and Jean{-}Pierre David.
\newblock Binaryconnect: Training deep neural networks with binary weights
  during propagations.
\newblock \emph{CoRR}, 2015.

\bibitem[Dai et~al.(2019)Dai, Yang, Yang, Cohen, Carbonell, Le, and
  Salakhutdinov]{dai2019transformer}
Zihang Dai, Zhilin Yang, Yiming Yang, William~W Cohen, Jaime Carbonell, Quoc~V
  Le, and Ruslan Salakhutdinov.
\newblock Transformer-xl: Attentive language models beyond a fixed-length
  context.
\newblock \emph{arXiv preprint arXiv:1901.02860}, 2019.

\bibitem[Dauphin et~al.(2017)Dauphin, Fan, Auli, and
  Grangier]{dauphin2017convlm}
Yann~N. Dauphin, Angela Fan, Michael Auli, and David Grangier.
\newblock Language modeling with gated convolutional networks.
\newblock In \emph{Proc. of ICML}, 2017.

\bibitem[Dehghani et~al.(2018)Dehghani, Gouws, Vinyals, Uszkoreit, and Łukasz
  Kaiser]{dehghani2018universal}
Mostafa Dehghani, Stephan Gouws, Oriol Vinyals, Jakob Uszkoreit, and Łukasz
  Kaiser.
\newblock Universal transformers, 2018.

\bibitem[Deng et~al.(2009)Deng, Dong, Socher, Li, Li, and
  Fei-Fei]{imagenet_cvpr09}
J.~Deng, W.~Dong, R.~Socher, L.-J. Li, K.~Li, and L.~Fei-Fei.
\newblock {ImageNet: A Large-Scale Hierarchical Image Database}.
\newblock In \emph{CVPR}, 2009.

\bibitem[Devlin et~al.(2018)Devlin, Chang, Lee, and Toutanova]{devlin2018bert}
Jacob Devlin, Ming-Wei Chang, Kenton Lee, and Kristina Toutanova.
\newblock Bert: Pre-training of deep bidirectional transformers for language
  understanding.
\newblock \emph{arXiv preprint arXiv:1810.04805}, 2018.

\bibitem[Dong et~al.(2019)Dong, Ni, Li, Chen, Su, and Zhu]{dong2019stochastic}
Yinpeng Dong, Renkun Ni, Jianguo Li, Yurong Chen, Hang Su, and Jun Zhu.
\newblock Stochastic quantization for learning accurate low-bit deep neural
  networks.
\newblock \emph{International Journal of Computer Vision}, 127\penalty0
  (11-12):\penalty0 1629--1642, 2019.

\bibitem[Fan et~al.(2019)Fan, Grave, and Joulin]{fan2019reducing}
Angela Fan, Edouard Grave, and Armand Joulin.
\newblock Reducing transformer depth on demand with structured dropout.
\newblock \emph{arXiv preprint arXiv:1909.11556}, 2019.

\bibitem[Gal \& Ghahramani(2016)Gal and Ghahramani]{gal2016dropout}
Yarin Gal and Zoubin Ghahramani.
\newblock Dropout as a bayesian approximation: Representing model uncertainty
  in deep learning.
\newblock In \emph{international conference on machine learning}, pp.\
  1050--1059, 2016.

\bibitem[Gong et~al.(2014)Gong, Liu, Yang, and Bourdev]{gong2014compressing}
Yunchao Gong, Liu Liu, Ming Yang, and Lubomir Bourdev.
\newblock Compressing deep convolutional networks using vector quantization.
\newblock \emph{arXiv preprint arXiv:1412.6115}, 2014.

\bibitem[Grave et~al.(2016)Grave, Joulin, Cisse, Grangier, and
  Jegou]{grave2016arxiv}
Edouard Grave, Armand Joulin, Moustapha Cisse, David Grangier, and Herve Jegou.
\newblock Efficient softmax approximation for gpus.
\newblock \emph{arXiv}, abs/1609.04309, 2016.

\bibitem[Gupta et~al.(2015)Gupta, Agrawal, Gopalakrishnan, and
  Narayanan]{gupta2015deep}
Suyog Gupta, Ankur Agrawal, Kailash Gopalakrishnan, and Pritish Narayanan.
\newblock Deep learning with limited numerical precision.
\newblock In \emph{ICML}, 2015.

\bibitem[Han et~al.(2015)Han, Pool, Tran, and Dally]{han2015learning}
Song Han, Jeff Pool, John Tran, and William Dally.
\newblock Learning both weights and connections for efficient neural network.
\newblock In \emph{NIPS}, pp.\  1135--1143, 2015.

\bibitem[Han et~al.(2016)Han, Mao, and Dally]{han2015deep}
Song Han, Huizi Mao, and William~J. Dally.
\newblock Deep compression: Compressing deep neural networks with pruning,
  trained quantization and {H}uffman coding.
\newblock \emph{ICLR}, 2016.

\bibitem[He et~al.(2015)He, Zhang, Ren, and Sun]{DBLP:journals/corr/HeZRS15}
Kaiming He, Xiangyu Zhang, Shaoqing Ren, and Jian Sun.
\newblock Deep residual learning for image recognition.
\newblock \emph{CoRR}, 2015.

\bibitem[Hinton et~al.(2015)Hinton, Vinyals, and Dean]{hinton2015distilling}
Geoffrey Hinton, Oriol Vinyals, and Jeff Dean.
\newblock Distilling the knowledge in a neural network.
\newblock \emph{arXiv preprint arXiv:1503.02531}, 2015.

\bibitem[Howard et~al.(2019)Howard, Sandler, Chu, Chen, Chen, Tan, Wang, Zhu,
  Pang, Vasudevan, Le, and Adam]{howard2019searching}
Andrew Howard, Mark Sandler, Grace Chu, Liang-Chieh Chen, Bo~Chen, Mingxing
  Tan, Weijun Wang, Yukun Zhu, Ruoming Pang, Vijay Vasudevan, Quoc~V. Le, and
  Hartwig Adam.
\newblock Searching for mobilenetv3.
\newblock \emph{arXiv e-prints}, 2019.

\bibitem[Huang et~al.(2016)Huang, Sun, Liu, Sedra, and
  Weinberger]{huang2016deep}
Gao Huang, Yu~Sun, Zhuang Liu, Daniel Sedra, and Kilian~Q Weinberger.
\newblock Deep networks with stochastic depth.
\newblock In \emph{ECCV}, 2016.

\bibitem[Huang et~al.(2018)Huang, Liu, Van~der Maaten, and
  Weinberger]{huang2018condensenet}
Gao Huang, Shichen Liu, Laurens Van~der Maaten, and Kilian~Q Weinberger.
\newblock Condensenet: An efficient densenet using learned group convolutions.
\newblock In \emph{CVPR}, 2018.

\bibitem[Jacob et~al.(2018)Jacob, Kligys, Chen, Zhu, Tang, Howard, Adam, and
  Kalenichenko]{jacob2018quantization}
Benoit Jacob, Skirmantas Kligys, Bo~Chen, Menglong Zhu, Matthew Tang, Andrew
  Howard, Hartwig Adam, and Dmitry Kalenichenko.
\newblock Quantization and training of neural networks for efficient
  integer-arithmetic-only inference.
\newblock In \emph{Proceedings of the IEEE Conference on Computer Vision and
  Pattern Recognition}, pp.\  2704--2713, 2018.

\bibitem[Jegou et~al.(2011)Jegou, Douze, and
  Schmid]{Jegou:2011:PQN:1916487.1916695}
Herve Jegou, Matthijs Douze, and Cordelia Schmid.
\newblock Product quantization for nearest neighbor search.
\newblock \emph{PAMI}, 2011.

\bibitem[Jiao et~al.(2019)Jiao, Yin, Shang, Jiang, Chen, Li, Wang, and
  Liu]{jiao2019tinybert}
Xiaoqi Jiao, Yichun Yin, Lifeng Shang, Xin Jiang, Xiao Chen, Linlin Li, Fang
  Wang, and Qun Liu.
\newblock Tinybert: Distilling bert for natural language understanding.
\newblock \emph{arXiv preprint arXiv:1909.10351}, 2019.

\bibitem[Joulin et~al.(2016)Joulin, Grave, Bojanowski, Douze, J{\'e}gou, and
  Mikolov]{joulin2016fasttext}
Armand Joulin, Edouard Grave, Piotr Bojanowski, Matthijs Douze, H{\'e}rve
  J{\'e}gou, and Tomas Mikolov.
\newblock Fasttext.zip: Compressing text classification models.
\newblock \emph{arXiv preprint arXiv:1612.03651}, 2016.

\bibitem[Krishnamoorthi(2018)]{krishnamoorthi2018quantizing}
Raghuraman Krishnamoorthi.
\newblock Quantizing deep convolutional networks for efficient inference: A
  whitepaper.
\newblock \emph{arXiv preprint arXiv:1806.08342}, 2018.

\bibitem[Lample \& Conneau(2019)Lample and Conneau]{lample2019cross}
Guillaume Lample and Alexis Conneau.
\newblock Cross-lingual language model pretraining.
\newblock \emph{arXiv preprint arXiv:1901.07291}, 2019.

\bibitem[Lan et~al.(2019)Lan, Chen, Goodman, Gimpel, Sharma, and
  Soricut]{lan2019albert}
Zhenzhong Lan, Mingda Chen, Sebastian Goodman, Kevin Gimpel, Piyush Sharma, and
  Radu Soricut.
\newblock Albert: A lite bert for self-supervised learning of language
  representations, 2019.

\bibitem[LeCun et~al.(1990)LeCun, Denker, and Solla]{lecun1990optimal}
Yann LeCun, John~S. Denker, and Sara~A. Solla.
\newblock Optimal brain damage.
\newblock In \emph{NIPS}, 1990.

\bibitem[Li et~al.(2016)Li, Kadav, Durdanovic, Samet, and Graf]{li2016pruning}
Hao Li, Asim Kadav, Igor Durdanovic, Hanan Samet, and Hans~Peter Graf.
\newblock Pruning filters for efficient convnets.
\newblock \emph{arXiv preprint arXiv:1608.08710}, 2016.

\bibitem[Li et~al.(2019)Li, Dong, and Wang]{li2019additive}
Yuhang Li, Xin Dong, and Wei Wang.
\newblock Additive powers-of-two quantization: A non-uniform discretization for
  neural networks.
\newblock \emph{arXiv preprint arXiv:1909.13144}, 2019.

\bibitem[Liu et~al.(2019)Liu, Ott, Goyal, Du, Joshi, Chen, Levy, Lewis,
  Zettlemoyer, and Stoyanov]{liu2019roberta}
Yinhan Liu, Myle Ott, Naman Goyal, Jingfei Du, Mandar Joshi, Danqi Chen, Omer
  Levy, Mike Lewis, Luke Zettlemoyer, and Veselin Stoyanov.
\newblock Roberta: A robustly optimized bert pretraining approach.
\newblock \emph{arXiv preprint arXiv:1907.11692}, 2019.

\bibitem[Liu et~al.(2018)Liu, Sun, Zhou, Huang, and Darrell]{liu2018rethinking}
Zhuang Liu, Mingjie Sun, Tinghui Zhou, Gao Huang, and Trevor Darrell.
\newblock Rethinking the value of network pruning.
\newblock \emph{arXiv preprint arXiv:1810.05270}, 2018.

\bibitem[Loshchilov \& Hutter(2016)Loshchilov and Hutter]{loshchilov2016sgdr}
Ilya Loshchilov and Frank Hutter.
\newblock Sgdr: Stochastic gradient descent with warm restarts.
\newblock \emph{arXiv preprint arXiv:1608.03983}, 2016.

\bibitem[Louizos et~al.(2017)Louizos, Welling, and Kingma]{louizos2017learning}
Christos Louizos, Max Welling, and Diederik~P Kingma.
\newblock Learning sparse neural networks through $ l\_0 $ regularization.
\newblock \emph{arXiv preprint arXiv:1712.01312}, 2017.

\bibitem[Luo et~al.(2017)Luo, Wu, and Lin]{luo2017thinet}
Jian-Hao Luo, Jianxin Wu, and Weiyao Lin.
\newblock Thinet: A filter level pruning method for deep neural network
  compression.
\newblock In \emph{ICCV}, 2017.

\bibitem[Ma et~al.(2018)Ma, Zhang, Zheng, and
  Sun]{DBLP:journals/corr/abs-1807-11164}
Ningning Ma, Xiangyu Zhang, Hai{-}Tao Zheng, and Jian Sun.
\newblock Shufflenet {V2:} practical guidelines for efficient {CNN}
  architecture design.
\newblock \emph{CoRR}, 2018.

\bibitem[Ma et~al.(2019)Ma, Zhang, Zhang, Duan, Hou, Song, and
  Zhou]{ma2019tensorized}
Xindian Ma, Peng Zhang, Shuai Zhang, Nan Duan, Yuexian Hou, Dawei Song, and
  Ming Zhou.
\newblock A tensorized transformer for language modeling.
\newblock \emph{arXiv preprint arXiv:1906.09777}, 2019.

\bibitem[McDonnell(2018)]{mcdonnell2018training}
Mark~D. McDonnell.
\newblock Training wide residual networks for deployment using a single bit for
  each weight, 2018.

\bibitem[Merity et~al.(2016)Merity, Xiong, Bradbury, and
  Socher]{merity2016wikitext}
Stephen Merity, Caiming Xiong, James Bradbury, and Richard Socher.
\newblock {Pointer Sentinel Mixture Models}.
\newblock \emph{arXiv}, abs/1609.07843, 2016.

\bibitem[Mittal et~al.(2018)Mittal, Bhardwaj, Khapra, and
  Ravindran]{mittal2018recovering}
Deepak Mittal, Shweta Bhardwaj, Mitesh~M Khapra, and Balaraman Ravindran.
\newblock Recovering from random pruning: On the plasticity of deep
  convolutional neural networks.
\newblock In \emph{WACV}, 2018.

\bibitem[Molchanov et~al.(2017)Molchanov, Ashukha, and
  Vetrov]{molchanov2017variational}
Dmitry Molchanov, Arsenii Ashukha, and Dmitry Vetrov.
\newblock Variational dropout sparsifies deep neural networks.
\newblock In \emph{ICML}, 2017.

\bibitem[Ott et~al.(2019)Ott, Edunov, Baevski, Fan, Gross, Ng, Grangier, and
  Auli]{ott2019fairseq}
Myle Ott, Sergey Edunov, Alexei Baevski, Angela Fan, Sam Gross, Nathan Ng,
  David Grangier, and Michael Auli.
\newblock fairseq: A fast, extensible toolkit for sequence modeling.
\newblock In \emph{Proceedings of NAACL-HLT 2019: Demonstrations}, 2019.

\bibitem[Pascanu et~al.(2014)Pascanu, Gulcehre, Cho, and
  Bengio]{pascanu2014construct}
Razvan Pascanu, Caglar Gulcehre, Kyunghyun Cho, and Yoshua Bengio.
\newblock How to construct deep recurrent neural networks.
\newblock In \emph{Proceedings of the Second International Conference on
  Learning Representations (ICLR 2014)}, 2014.

\bibitem[Paszke et~al.(2017)Paszke, Gross, Chintala, Chanan, Yang, DeVito, Lin,
  Desmaison, Antiga, and Lerer]{paszke2017automatic}
Adam Paszke, Sam Gross, Soumith Chintala, Gregory Chanan, Edward Yang, Zachary
  DeVito, Zeming Lin, Alban Desmaison, Luca Antiga, and Adam Lerer.
\newblock Automatic differentiation in pytorch.
\newblock 2017.

\bibitem[Radford et~al.(2019)Radford, Wu, Child, Luan, Amodei, and
  Sutskever]{radford2019language}
Alec Radford, Jeffrey Wu, Rewon Child, David Luan, Dario Amodei, and Ilya
  Sutskever.
\newblock Language models are unsupervised multitask learners.
\newblock 2019.

\bibitem[Rae et~al.(2019)Rae, Potapenko, Jayakumar, and
  Lillicrap]{rae2019compressive}
Jack~W Rae, Anna Potapenko, Siddhant~M Jayakumar, and Timothy~P Lillicrap.
\newblock Compressive transformers for long-range sequence modelling.
\newblock \emph{arXiv preprint arXiv:1911.05507}, 2019.

\bibitem[Rastegari et~al.(2016)Rastegari, Ordonez, Redmon, and
  Farhadi]{rastegari2016xnor}
Mohammad Rastegari, Vicente Ordonez, Joseph Redmon, and Ali Farhadi.
\newblock Xnor-net: Imagenet classification using binary convolutional neural
  networks.
\newblock In \emph{ECCV}, 2016.

\bibitem[Sandler et~al.(2018)Sandler, Howard, Zhu, Zhmoginov, and
  Chen]{sandler2018mobilenetv2}
Mark Sandler, Andrew Howard, Menglong Zhu, Andrey Zhmoginov, and Liang-Chieh
  Chen.
\newblock Mobilenetv2: Inverted residuals and linear bottlenecks.
\newblock In \emph{Conference on Computer Vision and Pattern Recognition}, pp.\
   4510--4520, 2018.

\bibitem[Sanh et~al.(2019{\natexlab{a}})Sanh, Debut, Chaumond, and
  Wolf]{distilbert}
Victor Sanh, Lysandre Debut, Julien Chaumond, and Thomas Wolf.
\newblock Distilbert, a distilled version of bert: smaller, faster, cheaper and
  lighter.
\newblock 2019{\natexlab{a}}.

\bibitem[Sanh et~al.(2019{\natexlab{b}})Sanh, Debut, Chaumond, and
  Wolf]{sanh2019distilbert}
Victor Sanh, Lysandre Debut, Julien Chaumond, and Thomas Wolf.
\newblock Distilbert, a distilled version of bert: smaller, faster, cheaper and
  lighter.
\newblock \emph{arXiv preprint arXiv:1910.01108}, 2019{\natexlab{b}}.

\bibitem[Srivastava et~al.(2015)Srivastava, Greff, and
  Schmidhuber]{srivastava2015highway}
Rupesh~Kumar Srivastava, Klaus Greff, and J{\"u}rgen Schmidhuber.
\newblock Highway networks.
\newblock \emph{arXiv preprint arXiv:1505.00387}, 2015.

\bibitem[Stock et~al.(2019)Stock, Joulin, Gribonval, Graham, and
  J{\'{e}}gou]{DBLP:journals/corr/abs-1907-05686}
Pierre Stock, Armand Joulin, R{\'{e}}mi Gribonval, Benjamin Graham, and
  Herv{\'{e}} J{\'{e}}gou.
\newblock And the bit goes down: Revisiting the quantization of neural
  networks.
\newblock \emph{CoRR}, abs/1907.05686, 2019.

\bibitem[Sukhbaatar et~al.(2019{\natexlab{a}})Sukhbaatar, Grave, Bojanowski,
  and Joulin]{sukhbaatar2019adaptive}
Sainbayar Sukhbaatar, Edouard Grave, Piotr Bojanowski, and Armand Joulin.
\newblock Adaptive attention span in transformers.
\newblock \emph{arXiv preprint arXiv:1905.07799}, 2019{\natexlab{a}}.

\bibitem[Sukhbaatar et~al.(2019{\natexlab{b}})Sukhbaatar, Grave, Lample, Jegou,
  and Joulin]{sukhbaatar2019augmenting}
Sainbayar Sukhbaatar, Edouard Grave, Guillaume Lample, Herve Jegou, and Armand
  Joulin.
\newblock Augmenting self-attention with persistent memory.
\newblock \emph{arXiv preprint arXiv:1907.01470}, 2019{\natexlab{b}}.

\bibitem[Sun et~al.(2019)Sun, Cheng, Gan, and Liu]{Sun_2019}
Siqi Sun, Yu~Cheng, Zhe Gan, and Jingjing Liu.
\newblock Patient knowledge distillation for bert model compression.
\newblock \emph{EMNLP}, 2019.

\bibitem[Sun et~al.()Sun, Yu, Song, Liu, Yang, and Zhou]{sunmobilebert}
Zhiqing Sun, Hongkun Yu, Xiaodan Song, Renjie Liu, Yiming Yang, and Denny Zhou.
\newblock Mobilebert: Task-agnostic compression of bert for resource limited
  devices.

\bibitem[Sutskever et~al.(2013)Sutskever, Martens, Dahl, and
  Hinton]{sutskever2013importance}
Ilya Sutskever, James Martens, George Dahl, and Geoffrey Hinton.
\newblock On the importance of initialization and momentum in deep learning.
\newblock In \emph{International conference on machine learning}, pp.\
  1139--1147, 2013.

\bibitem[Tan \& Le(2019)Tan and Le]{tan2019efficientnet}
Mingxing Tan and Quoc~V. Le.
\newblock Efficientnet: Rethinking model scaling for convolutional neural
  networks, 2019.

\bibitem[Turc et~al.(2019)Turc, Chang, Lee, and Toutanova]{turc2019well}
Iulia Turc, Ming-Wei Chang, Kenton Lee, and Kristina Toutanova.
\newblock Well-read students learn better: The impact of student initialization
  on knowledge distillation.
\newblock \emph{arXiv preprint arXiv:1908.08962}, 2019.

\bibitem[Vanhoucke et~al.(2011)Vanhoucke, Senior, and
  Mao]{vanhoucke2011improving}
Vincent Vanhoucke, Andrew Senior, and Mark~Z Mao.
\newblock Improving the speed of neural networks on cpus.
\newblock 2011.

\bibitem[Vaswani et~al.(2017)Vaswani, Shazeer, Parmar, Uszkoreit, Jones, Gomez,
  Kaiser, and Polosukhin]{vaswani2017attention}
Ashish Vaswani, Noam Shazeer, Niki Parmar, Jakob Uszkoreit, Llion Jones,
  Aidan~N Gomez, {\L}ukasz Kaiser, and Illia Polosukhin.
\newblock Attention is all you need.
\newblock In \emph{NIPS}, 2017.

\bibitem[Wan et~al.(2013)Wan, Zeiler, Zhang, Le~Cun, and
  Fergus]{wan2013regularization}
Li~Wan, Matthew Zeiler, Sixin Zhang, Yann Le~Cun, and Rob Fergus.
\newblock Regularization of neural networks using {D}rop{C}onnect.
\newblock In \emph{ICML}, 2013.

\bibitem[Wang et~al.(2019)Wang, Singh, Michael, Hill, Levy, and
  Bowman]{wang2019glue}
Alex Wang, Amanpreet Singh, Julian Michael, Felix Hill, Omer Levy, and
  Samuel~R. Bowman.
\newblock {GLUE}: A multi-task benchmark and analysis platform for natural
  language understanding.
\newblock 2019.
\newblock ICLR.

\bibitem[Wang et~al.(2018)Wang, Liu, andx Ji~Lin, and
  Han]{DBLP:journals/corr/abs-1811-08886}
Kuan Wang, Zhijian Liu, Yujun~Lin andx Ji~Lin, and Song Han.
\newblock {HAQ:} hardware-aware automated quantization.
\newblock \emph{CoRR}, 2018.

\bibitem[Williams et~al.(2018)Williams, Nangia, and Bowman]{williams2018broad}
Adina Williams, Nikita Nangia, and Samuel~R. Bowman.
\newblock A broad-coverage challenge corpus for sentence understanding through
  inference.
\newblock In \emph{Proceedings of NAACL-HLT}, 2018.

\bibitem[Wu et~al.(2019)Wu, Fan, Baevski, Dauphin, and Auli]{wu2018pay}
Felix Wu, Angela Fan, Alexei Baevski, Yann Dauphin, and Michael Auli.
\newblock Pay less attention with lightweight and dynamic convolutions.
\newblock In \emph{ICLR}, 2019.

\bibitem[Zhang et~al.(2018)Zhang, Xiong, and Su]{zhang2018accelerating}
Biao Zhang, Deyi Xiong, and Jinsong Su.
\newblock Accelerating neural transformer via an average attention network.
\newblock \emph{arXiv preprint arXiv:1805.00631}, 2018.

\bibitem[Zhang et~al.(2017)Zhang, Zhou, Lin, and
  Sun]{DBLP:journals/corr/ZhangZLS17}
Xiangyu Zhang, Xinyu Zhou, Mengxiao Lin, and Jian Sun.
\newblock Shufflenet: An extremely efficient convolutional neural network for
  mobile devices.
\newblock \emph{CoRR}, 2017.

\bibitem[Zhao et~al.(2019)Zhao, Gupta, Song, and Zhou]{zhao2019extreme}
Sanqiang Zhao, Raghav Gupta, Yang Song, and Denny Zhou.
\newblock Extreme language model compression with optimal subwords and shared
  projections.
\newblock \emph{arXiv preprint arXiv:1909.11687}, 2019.

\end{thebibliography}
\bibliographystyle{iclr2021_conference}

\newpage
\clearpage
\section{Appendix}

\subsection{Quantization of Additional Architectures}

\begin{table}[t]
\centering 
    
    \begin{tabular}{llcc}
    \toprule
    \bf Setting & \bf Model & \bf Compression & \bf Top-1 Accuracy\\
    \midrule 
    \multirow{2}{*}{Small Blocks} & \cite{DBLP:journals/corr/abs-1907-05686} & 19x & 73.8 \\ 
    & Quant-Noise & 19x & \bf 74.3\\ 
    \midrule 
    \multirow{2}{*}{Large Blocks} & \cite{DBLP:journals/corr/abs-1907-05686} & 32x & 68.2 \\ 
    & Quant-Noise & 32x & \bf 68.8 \\ 
    \bottomrule
    \end{tabular}
    \caption{    
\textbf{Compression of ResNet-50 with Quant-Noise}. We compare to \cite{DBLP:journals/corr/abs-1907-05686} in both the small and large blocks regime. For fair comparison, we hold the compression rate constant. Quant-Noise provides improved performance in both settings. 
}
    \label{tab:resnet}
\end{table}

\paragraph{ResNet-50.} We explore the compression of ResNet-50, a standard architecture used Computer Vision. In Table~\ref{tab:resnet}, we compare Quant-Noise to iPQ Compression from \cite{DBLP:journals/corr/abs-1907-05686} and show that Quant-Noise provide consistent additional improvement.

\subsection{Ablations} 
\label{sec:ablations}

In this section, we examine the impact of the level of noise during training  as well as the impact of approximating iPQ during training. 

\begin{figure}[t]
    \includegraphics[width=.48\textwidth]{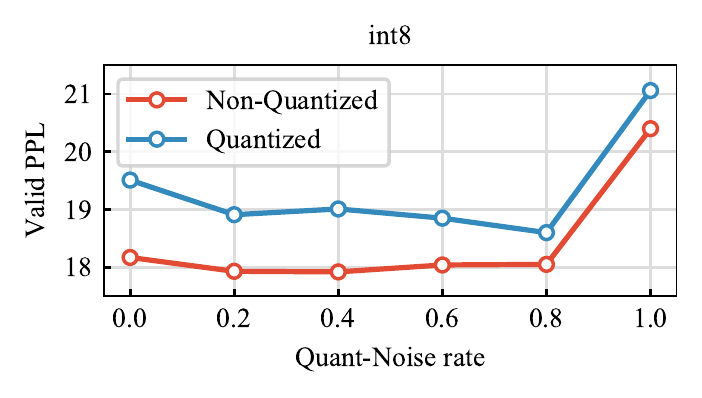}     \hfill
    \includegraphics[width=.48\textwidth]{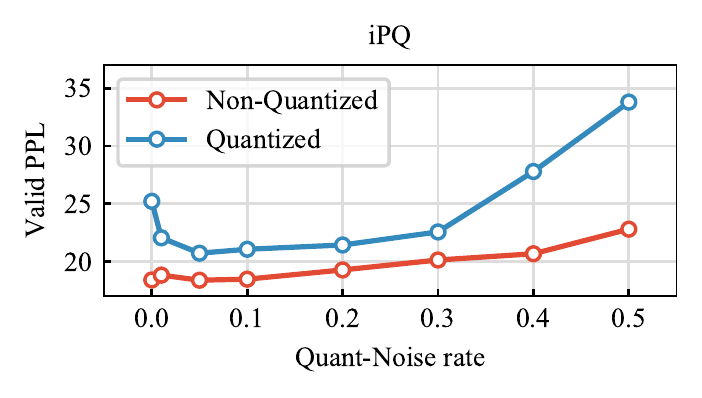} \vspace{-5pt} 
    
    \captionof{figure}{\textbf{Effect of Quantization Parameters}. We report the influence of the proportion of blocks to which we apply the noise. We focus on Transformer for Wikitext-103 language modeling. We explore two settings: iPQ and \texttt{int8}. For iPQ, we use $\varphi_{\text{proxy}}$.}
    \label{fig:ablation_noise}
\end{figure}

\begin{table}[t]
\begin{minipage}[t][][b]{0.48\linewidth}
    \begin{tabular}{llcc}
    \toprule
    \bf Noise & \bf Blocks & \bf PPL & \bf Quant PPL\\
    \midrule 
    $\varphi_{\text{PQ}}$ & Subvectors & 18.3 & 21.1 \\ 
    $\varphi_{\text{PQ}}$ & Clusters & 18.3 & 21.2 \\ 
    \midrule
    $\varphi_{\text{proxy}}$ & Subvectors & 18.3 & 21.0 \\ 
    $\varphi_{\text{proxy}}$ & Clusters & 18.4 & 21.1 \\ 
    \bottomrule
    \end{tabular}
\end{minipage}
\begin{minipage}[t][][b]{0.48\linewidth}
    \caption{
\textbf{Exact versus proxy noise function for different block selections with iPQ.} 
We compare exact $\phi_{\text{PQ}}$ and the approximation $\phi_{\text{proxy}}$ with blocks selected from all subvectors or subvectors from the same cluster.
}
    \label{tab:wikitext103_noise}
\end{minipage}
\end{table}

\subsection{Impact of Noise Rate}

We analyze the performance for various values of \method in Figure~\ref{fig:ablation_noise} on a Transformer for language modeling.
For iPQ, performance is impacted by high rates of quantization noise.
For example, a Transformer with the noise function $\varphi_{\text{proxy}}$ degrades with rate higher than 0.5, i.e., when half of the weights are passed through the noise function $\varphi_{\text{proxy}}$.
We hypothesize that for large quantities of noise, a larger effect of using proxy rather than the exact PQ noise is observed.
For \texttt{int8} quantization and its noise function, higher rates of noise are slightly worse but not as severe.
A rate of $1$ for \texttt{int8} quantization is equivalent to the Quantization Aware Training of~\citep{krishnamoorthi2018quantizing}, as the full matrix is quantized with STE, showing the potential benefit of partial quantization during training.

\subsection{Impact of Approximating the Noise Function}

We study the impact of approximating quantization noise during training. 
We focus on the case of iPQ with the approximation described in Section~\ref{sect:ipq_application}.
In Table~\ref{tab:wikitext103_noise}, we compare the correct noise function for iPQ with its approximation $\varphi_{\text{proxy}}$.
This approximate noise function does not consider cluster assignments or centroid values and simply zeroes out the selected blocks.
For completeness, we include an intermediate approximation where we consider cluster assignments to apply noise within each cluster, but still zero-out the vectors.
These approximations do not affect the performance of the quantized models.
This suggests that increasing the correlation between subvectors that are jointly clustered is enough to maintain the performance of a model quantized with iPQ.
Since PQ tends to work well on highly correlated vectors, such as activations in convolutional networks, this is not surprising. Using the approximation $\varphi_{\text{proxy}}$ presents the advantage of speed and practicality. Indeed, one does not need to compute cluster assignments and centroids for every layer in the network after each epoch. Moreover, the approach $\varphi_{\text{proxy}}$ is less involved in terms of code.

\subsection{Experimental Setting}
\label{sec:experiments}

We assess the effectiveness of \method on competitive language and vision benchmarks.
We consider Transformers for language modeling, RoBERTa for pre-training sentence representations, and EfficientNet for image classification. 
Our models are implemented in PyTorch \citep{paszke2017automatic}. 
We use \texttt{fairseq} \citep{ott2019fairseq} for language modeling and pre-training for sentence representation tasks and~\texttt{Classy Vision}~\citep{adcock2019classy} for EfficientNet.

\paragraph{Language Modeling.} 
We experiment on the \texttt{Wikitext-103} benchmark~\citep{merity2016wikitext} that contains $100$M tokens and a vocabulary of $260$k words.
We train a~$16$ layer Transformer following~\cite{baevski2018adaptive} with a LayerDrop rate of 0.2~\citep{fan2019reducing}. 
We report perplexity (PPL) on the test set. 
\vspace{-0.5em}
\paragraph{Pre-Training of Sentence Representations.} 
We pre-train the base BERT model~\citep{devlin2018bert} on the \texttt{BooksCorpus + Wiki} dataset with a LayerDrop rate of $0.2$. 
We finetune the pre-trained models on the MNLI task~\citep{williams2018broad} from the GLUE Benchmark~\citep{wang2019glue} and report accuracy. 
We follow the parameters in~\cite{liu2019roberta} training and finetuning.
\vspace{-0.5em}
\paragraph{Image Classification.}
We train an EfficientNet-B3 model~\citep{tan2019efficientnet} on the ImageNet object classification benchmark \citep{imagenet_cvpr09}.
The EfficientNet-B3 of \texttt{Classy Vision} achieves a Top-1 accuracy of $81.5\%$, which is slightly below than the performance of $81.9\%$ reported by~\cite{tan2019efficientnet}.

\subsection{Training Details} 

\paragraph{Language Modeling} 

To handle the large vocabulary of Wikitext-103, we follow \citep{dauphin2017convlm} and \citep{baevski2018adaptive} in using adaptive softmax \citep{grave2016arxiv} and adaptive input for computational efficiency. For both input and output embeddings, we use dimension size $1024$ and three adaptive bands: $20$K, $40$K, and $200$K. We use a cosine learning rate schedule \citep{baevski2018adaptive,loshchilov2016sgdr} and train with Nesterov's accelerated gradient \citep{sutskever2013importance}. We set the momentum to 0.99 and renormalize gradients if the norm exceeds 0.1 \citep{pascanu2014construct}. During training, we partition the data into blocks of contiguous tokens that ignore document boundaries. At test time, we respect sentence boundaries. We set LayerDrop to 0.2. We set \method value to 0.05. During training time, we searched over the parameters (0.05, 0.1, 0.2) to determine the optimal value of \method. During training time, the block size of \method is 8.

\paragraph{RoBERTa}

The base architecture is a $12$ layer model with embedding size $768$ and FFN size $3072$. We follow \citep{liu2019roberta} in using the subword tokenization scheme from \citep{radford2019language}, which uses bytes as subword units. This eliminates unknown tokens. We train with large batches of size $8192$ and maintain this batch size using gradient accumulation. We do not use next sentence prediction~\citep{lample2019cross}. We optimize with Adam with a polynomial decay learning rate schedule. We set LayerDrop to 0.2. We set \method value to 0.1. We did not hyperparameter search to determine the optimal value of \method as training RoBERTa is computationally intensive. During training time, the block size of \method is 8.

During finetuning, we hyperparameter search over three learning rate options (1e-5, 2e-5, 3e-5) and batchsize (16 or 32 sentences). The other parameters are set following \citep{liu2019roberta}. We do single task finetuning, meaning we only tune on the data provided for the given natural language understanding task. We do not perform ensembling. When finetuning models trained with LayerDrop, we apply LayerDrop and \method during finetuning time as well. 

\paragraph{EfficientNet}

We use the architecture of EfficientNet-B3 defined in \texttt{Classy Vision} \citep{adcock2019classy} and follow the default hyperparameters for training. We set \method value to 0.1. During training time, we searched over the parameters (0.05, 0.1, 0.2) to determine the optimal value of \method. During training time, the block size of \method is set to $4$ for all $1 \times 1$ convolutions, $9$ for depth-wise $3\times 3$ convolutions, $5$ for depth-wise $5 \times 5$ convolutions and $4$ for the classifier. For sharing, we shared weights between blocks 9-10, 11-12, 14-15, 16-17, 19-20-21, 22-23 and refer to blocks that share the same weights as a \emph{chunk}. For LayerDrop, we drop the chunks of blocks defined previously with probability 0.2 and evaluate only with chunks 9-10, 14-15 and 19-20-21.

\subsection{Scalar Quantization Details}

We closely follow the methodology of PyTorch 1.4. We emulate scalar quantization by quantizing the weights and the activations. The scales and zero points of activations are determined by doing a few forward passes ahead of the evaluation and then fixed. We use the \texttt{Histogram} method to compute $s$ and $z$, which aims at approximately minimizing the $L_2$ quantization error by adjusting $s$ and $z$. This scheme is a refinement of the \texttt{MinMax} scheme. Per channel quantization is also discussed in Table~\ref{tab:hist_channel_ablation}.

\begin{table}[t]
    \small
    \centering
    \begin{tabular}{lccc}
    \toprule
    \bf Model & \bf MB & \bf PPL\\
    \midrule
    Trans XL Large \citep{dai2019transformer} & 970 & 18.3 \\ 
    Compressive Trans \citep{rae2019compressive} & 970 & 17.1 \\ 
    GCNN \citep{dauphin2017convlm} & 870 & 37.2 \\ 
    4 Layer QRNN \citep{bradbury2016quasi} & 575 & 33.0 \\ 
    Trans XL Base \citep{dai2019transformer} & 570 & 24.0 \\ 
    Persis Mem \citep{sukhbaatar2019augmenting} & 506 & 20.6 \\ 
    Tensorized core-2 \citep{ma2019tensorized} & 325 & 18.9 \\ 
    \midrule 
    \method & \bf 38 & 20.7 \\ 
    \method + Share + Prune & 10 & 24.2 \\ 
    \bottomrule
    \end{tabular}
    \vspace{0.2cm}
    \caption{\textbf{Performance on Wikitext-103.} We report test set perplexity and model size in megabytes. Lower perplexity is better. 
    }
    \label{tab:lm_table}
\end{table}

\begin{table}[t]
    \small
    \centering
    \begin{tabular}{lccc}
    \toprule
    \bf Model & \bf MB & \bf MNLI \\
    \midrule
    RoBERTa Base + LD \citep{fan2019reducing} & 480 & 84.8 \\ 
    BERT Base \citep{devlin2018bert} & 420 & 84.4 \\ 
    PreTrained Distil \citep{turc2019well} & 257 & 82.5 \\ 
    DistilBERT \citep{sanh2019distilbert} & 250 & 81.8 \\
    MobileBERT* \citep{sunmobilebert} & 96 & 84.4 \\ 
    TinyBERT$\dagger$ \citep{jiao2019tinybert} & 55 & 82.8 \\ 
    ALBERT Base \citep{lan2019albert} & 45 & 81.6 \\ 
    AdaBERT$\dagger$ \citep{chen2020adabert} & 36 & 81.6 \\ 
    \midrule 
    \method & 38 & 83.6 \\ 
    \method + Share + Prune & 14 & 82.5 \\ 
    \bottomrule
    \end{tabular}
    \vspace{0.2cm}
    \caption{\textbf{Performance on MNLI.} We report accuracy and size in megabytes. * indicates distillation using BERT Large. $\dagger$ indicates training with data augmentation. Work from~\cite{Sun_2019} and~\cite{zhao2019extreme} do not report results on the dev set. \cite{decompbert} do not report model size. Higher accuracy is better.
    }
    \label{tab:bert_table}
\end{table}

\begin{table}[t]
    \small
    \centering
    \begin{tabular}{lccc}
    \toprule
    \bf Model & \bf MB & \bf Acc. \\
    \midrule
    EfficientNet-B7 \citep{tan2019efficientnet} & 260 & 84.4 \\
    ResNet-50 \citep{DBLP:journals/corr/HeZRS15} & 97.5 & 76.1 \\
    DenseNet-169 \citep{huang2018condensenet} & 53.4 & 76.2 \\
    EfficientNet-B0 \citep{tan2019efficientnet} & 20.2 & 77.3 \\
    MobileNet-v2 \citep{sandler2018mobilenetv2} & 13.4 & 71.9 \\
    Shufflenet-v2 $\times 1$ \citep{DBLP:journals/corr/abs-1807-11164} & 8.7 & 69.4\\
    \midrule
    HAQ 4 bits \citep{DBLP:journals/corr/abs-1811-08886} & 12.4 & 76.2 \\
    iPQ ResNet-50 \citep{DBLP:journals/corr/abs-1907-05686} & 5.09 & 76.1 \\
    \midrule 
    \method & 3.3 & 80.0 \\ 
    \method + Share + Prune & 2.3 & 77.8 \\ 
    \bottomrule
    \end{tabular}
    \vspace{0.2cm}
    \caption{\textbf{Performance on ImageNet.} We report accuracy and size in megabytes. Higher accuracy is better.
    }
    \label{tab:imagenet_table}
\end{table}

\subsection{iPQ Quantization Details}

\paragraph{Language Modeling}

We quantize FFN with block size 8, embeddings with block size 8, and attention with block size 4. We tuned the block size for attention between the values (4, 8) to find the best performance. Note that during training with apply \method to all the layers.

\paragraph{RoBERTa}

We quantize FFN with block size 4, embeddings with block size 4, and attention with block size 4. We tuned the block size between the values (4, 8) to find the best performance. Note that during training with apply \method to all the layers.

\paragraph{EfficientNet}

We quantize blocks sequentially and end up with the classifier. The block sizes are $4$ for all $1 \times 1$ convolutions, $9$ for depth-wise $3\times 3$ convolutions, $5$ for depth-wise $5 \times 5$ convolutions and $4$ for the classifier. Note that during training with apply \method to all the weights in InvertedResidual Blocks (except the Squeeze-Excitation subblocks), the head convolution and the classifier.

\subsection{Details of Pruning and Layer Sharing}

We apply the \textit{Every Other Layer} strategy from~\cite{fan2019reducing}. When combining layer sharing with pruning, we train models with shared layers and then prune chunks of shared layers. When sharing layers, the weights of adjacent layers are shared in chunks of two. For a concrete example, imagine we have a model with layers A, B, C, D, E, F, G, H. We share layers A and B, C and D, E and F, G and H. To prune, every other chunk would be pruned away, for example we could prune A, B, E, F.

\subsection{Numerical Results for Graphical Diagrams} 

We report the numerical values displayed in Figures~\ref{fig:main_result} in Table~\ref{tab:lm_table} for language modeling, Table~\ref{tab:bert_table} for BERT, and Table~\ref{tab:imagenet_table} for ImageNet.

\subsection{Further Ablations}

\subsubsection{Impact of \method for the Vision setup}

We provide another study showing the impact of the proportion of elements on which to apply \method in Table~\ref{tab:rate_imagenet}.

\begin{table}[t]
    \small
    \centering
    \begin{tabular}{l|cccccc}
    \toprule
    $p$ & 0 & 0.2 & 0.4 & 0.6 & 0.8 & 1 \\
    \midrule
    Top-1       & 80.66 & 80.83 & 80.82 & 80.88 & 80.92 & 80.64\\
    \bottomrule
    \end{tabular}
    \vspace{0.2cm}
    \caption{\textbf{Effect of Quantization Parameters}. We report the influence of the \method rate $p$ with Scalar Quantization (\texttt{int8}). We focus on EfficientNet for ImageNet classification.}
    \label{tab:rate_imagenet}
\end{table}

\begin{figure}[t]
    \centering
    \includegraphics[width=0.3\textwidth]{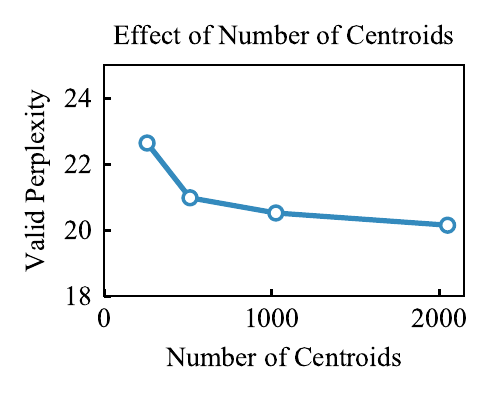}
    \caption{\textbf{Quantizing with a larger number of centroids}. Results are shown on Wikitext-103 valid.
}
    \label{fig:ablation_centroids}
\end{figure}

\begin{figure}[t]
    \centering
    \includegraphics[width=0.7\textwidth]{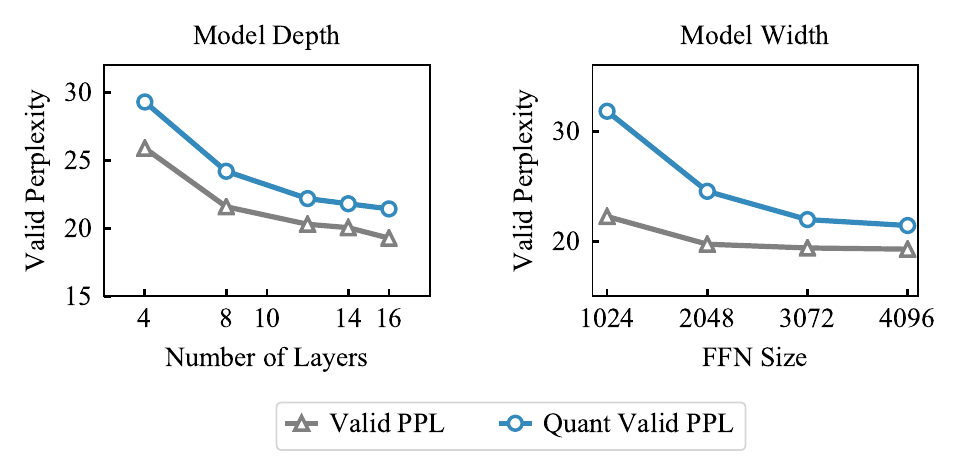}
    \caption{\textbf{(a)} Effect of Initial Model Size for more shallow models \textbf{(b)} Effect of Initial Model Size more skinny models
}
    \label{fig:ablation_appendix}
\end{figure}

\begin{figure}[t]
    \centering
    \includegraphics[width=0.7\textwidth]{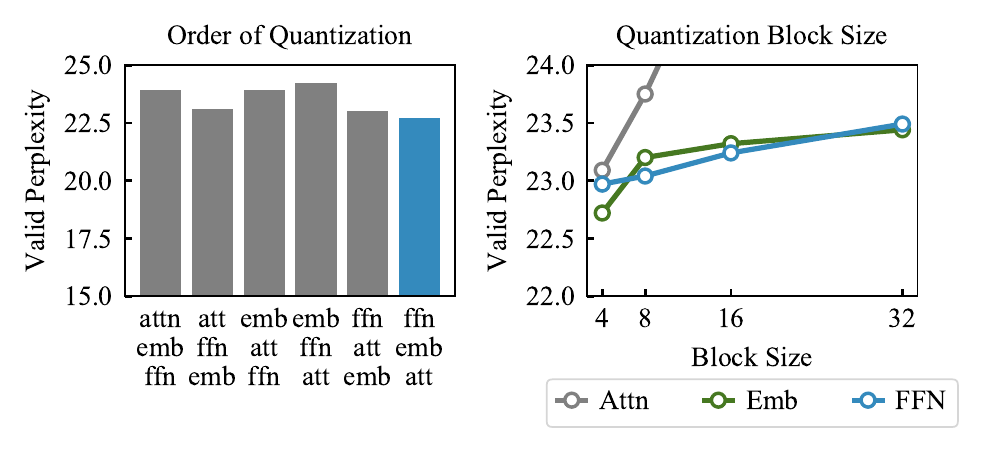}
    \caption{\textbf{Effect of Quantization on Model Structures.} Results are shown on the validation set of Wikitext-103. \textbf{(a)} Quantizing Attention, FFN, and Embeddings in different order. \textbf{(b)} More Extreme compression of different structures.
}
    \label{fig:ablation_3}
\end{figure}

\begin{table*}[th]
    \small
    \centering
    \begin{tabular}{l c ccc c ccc}
    \toprule
 \bf Quantization Scheme &~~& \multicolumn{3}{c}{\bf Language Modeling} &~~& \multicolumn{3}{c}{\bf Image Classification}\\
 && \multicolumn{3}{c}{\small 16-layer Transformer} &~~& \multicolumn{3}{c}{\small EfficientNet-B3}\\
 && \multicolumn{3}{c}{\small Wikitext-103} &~~& \multicolumn{3}{c}{\small ImageNet-1K}\\
\cmidrule{3-5}
\cmidrule{7-9}
&& Size  & Compress & Test PPL && Size & Compress & Top-1 Acc.\\
    \midrule
Uncompressed model && $942$ & $\times 1$ & $18.3$ && $46.7$ & $\times 1$ & $81.5$ \\
    \midrule
    Int4 Quant Histogram & & $118$ & $\times 8$ & $39.4$ && $5.8$ & $\times 8$ & $45.3$ \\ 
    + \method & & $118$ & $\times 8$ & $21.8$ && $5.8$ & $\times 8$\ & $67.8$  \\
    \rule{0pt}{3ex}Int4 Quant Channel & & $118$ & $\times 8$ & $21.2$&& $5.8$ & $\times 8$ & $68.2$\\ 
    + \method & & $118$ & $\times 8$ &  $19.5$ && $5.8$ & $\times 8$\ & $72.3$ \\
    \midrule 
    Int8 Quant Histogram & & $236$ & $\times 4$ & $19.6$ && $11.7$ & $\times 4$ & $80.7$\\ 
    + \method & &  $236$ & $\times 4$ & $18.7$  && $11.7$ & $\times 4$ & $80.9$\\
    \rule{0pt}{3ex} Int8 Quant Channel & & $236$ & $\times 4$ & $18.5$ &&  $11.7$ & $\times 4$ & $81.1$ \\ 
    + \method & &  $236$ & $\times 4$ & $18.3$  && $11.7$ & $\times 4$ & $81.2$\\
    \bottomrule
    \end{tabular}
    \caption{\textbf{Comparison of different approaches to \texttt{int4} and \texttt{int8} with and without \method} on language modeling and image classification. For language modeling, we train a Transformer on the Wikitext-103 benchmark. We report perplexity (PPL) on the test set. For image classification, we train a EfficientNet-B3 on the ImageNet-1K benchmark. We report top-1 accuracy on the validation set. For both setting, we also report model size in megabyte (MB) and the compression ratio compared to the original model.}
    \label{tab:hist_channel_ablation}
\end{table*}

\begin{table}
    \small
    \centering
    \begin{tabular}{lccc}
    \toprule
    \bf Model & \bf MB & \bf PPL\\
    \midrule
    \method + Share + Prune  & 10 & 24.2 \\ 
    \method + Share + Prune with STE & 10 & 24.5 \\ 
    \bottomrule
    \end{tabular}
    \vspace{0.2cm}
    \caption{\textbf{Performance on Wikitext-103 when using STE in the backward pass of the LayerDrop pruning noise.} 
    }
    \label{tab:prune_ste}
\end{table}

\subsubsection{Impact of the number of centroids} 
We quantize with 256 centroids which represents a balance between size and representation capacity. 
The effect of the number of centroids on performance and size is shown in  Figure~\ref{fig:ablation_centroids} (a).
Quantizing with more centroids improves perplexity --- this parameter could be adjusted based on the practical storage constraints. 

\subsubsection{Effect of Initial Model Size}

Large, overparameterized models are more easily compressed.
In Figure~\ref{fig:ablation_appendix}, we explore quantizing both shallower and skinnier models.
For shallow models, the gap between quantized and non-quantized perplexity does not increase as layers are removed (Figure~\ref{fig:ablation_appendix}, left). 
In contrast, there is a larger gap in performance for models with smaller FFN (Figure~\ref{fig:ablation_appendix}, right). 
As the FFN size decreases, the weights are less redundant and more difficult to quantize with iPQ.  

\subsubsection{Difficulty of Quantizing Different Model Structures}
Quantization is applied to various portions of the Transformer architecture --- the embedding, attention, feedforward, and classifier output. 
We compare the quantizability of various portions of the network in this section.

\paragraph{Is the order of structures important?}

We quantize specific network structures first --- this is important as quantizing weight matrices can accumulate reconstruction error.
Some structures of the network should be quantized last so the finetuning process can better adjust the centroids.
We find that there are small variations in performance based on quantization order (see Figure~\ref{fig:ablation_3}). We choose to quantize FFN, then embeddings, and finally the attention matrices in Transformer networks.

\paragraph{Which structures can be compressed the most?}

Finally, we analyze which network structures can be most compressed.
During quantization, various matrix block sizes can be chosen as a parameter --- the larger the block size, the more compression, but also the larger the potential reduction of performance.
Thus, it is important to understand how much each network structure can be compressed to reduce the memory footprint of the final model as much as possible.
In Figure~\ref{fig:ablation_3}, we quantize two model structures with a fixed block size and vary the block size of the third between 4 and 32.
As shown, the FFN and embedding structures are more robust to aggressive compression, while the attention drastically loses performance as larger block sizes are used. 

\subsubsection{Approach to \texttt{intN} Scalar Quantization}

We compare quantizing per-channel to using a histogram quantizer in Table~\ref{tab:hist_channel_ablation}. The histogram quantizer maintains a running min/max and minimizes L2 distance between quantized and non-quantized values to find the optimal min/max. Quantizing per channel learns scales and offsets as vectors along the channel dimension, which provides more flexibility since scales and offsets can be different. 

\subsubsection{LayerDrop with STE}

For quantization noise, we apply the straight through estimator (STE) to remaining weights in the backward pass. We experiment with applying STE to the backward pass of LayerDrop's pruning noise. Results are shown in Table~\ref{tab:prune_ste} and find slightly worse results.

\end{document}